\pdfoutput=1
\documentclass[11pt]{article}

\usepackage[final]{acl}

\usepackage{times}
\usepackage{latexsym}
\usepackage{times}
\usepackage{latexsym}
\usepackage{booktabs}
\usepackage{boxedminipage}
\usepackage{amssymb}
\usepackage{amsmath}
\usepackage{todonotes} 
\usepackage{graphicx}

\usepackage[T1]{fontenc}
\usepackage[utf8]{inputenc}

\usepackage{microtype}

\usepackage{hyperref}
\usepackage{inconsolata}
\usepackage{pifont}% http://ctan.org/pkg/pifont

\usepackage{tabu}
\usepackage{fontawesome}
\usepackage{amsmath}
\usepackage{amsfonts}
\usepackage{amssymb}
\usepackage{enumitem}
\usepackage{listings}
\usepackage{xstring}
\usepackage{graphicx}
\usepackage{pbox}
\usepackage{subcaption}
\usepackage{epstopdf}
\usepackage{xstring}
\usepackage{multirow}

\usepackage{soul}
\usepackage{color}
\usepackage{graphicx}
\usepackage{multirow}
\usepackage{comment}
\usepackage{listings} % Add this for using the listings package
\usepackage{graphicx}
\usepackage{subcaption}
\usepackage{verbatim}

\usepackage{arabtex}
\usepackage{utf8}
\setcode{utf8}

\usepackage{xcolor} % For coloring text
\definecolor{lightblue}{rgb}{.50,.90,0.51}
\definecolor{tri}{rgb}{.25,.88,.82}
\definecolor{lilac}{rgb}{0.85,0.64,0.85}
\definecolor{atomictangerine}{rgb}{1.0, 0.6, 0.4}

\title{ArMeme: Propagandistic Content in Arabic Memes\\\hphantom{...}
\\\footnotesize{\textcolor{red}{WARNING: This paper contains examples which may be disturbing to the reader}}}

\author{Firoj Alam$^1$, Abul Hasnat$^2$$^,$$^3$, Fatema Ahmad$^1$, Md Arid Hasan$^4$, Maram Hasanain$^1$\\
  $^1$Qatar Computing Research Institute, HBKU, Qatar\\  
  $^2$Blackbird.AI, USA, 
  $^3$APAVI.AI, France \\
  $^4$University of New Brunswick, Fredericton, NB, Canada \\
  {\tt \{fialam,fakter,mhasanain\}@hbku.edu.qa}, mhasnat@gmail.com, arid.hasan@unb.ca \\
\\}

\begin{document}
\maketitle

\begin{abstract}
With the rise of digital communication memes have become a significant medium for cultural and political expression that is often used to mislead audience. Identification of such misleading and persuasive multimodal content become more important among various stakeholders, including social media platforms, policymakers, and the broader society as they often cause harm to the individuals, organizations and/or society. While there has been effort to develop AI based automatic system for resource rich languages (e.g., English), it is relatively little to none for medium to low resource languages. In this study, we focused on developing an Arabic memes dataset with manual annotations of propagandistic content.\footnote{Propaganda is a form of communication designed to influence people's opinions or actions toward a specific goal, employing well-defined rhetorical and psychological techniques~\cite{InstituteforPropagandaAnalysis1938}.} We annotated $\sim6K$ Arabic memes collected from various social media platforms, which is a first resource for Arabic multimodal research. We provide a comprehensive analysis aiming to develop computational tools for their detection. We made the dataset publicly available for the community. 
\end{list}
\end{abstract}

% \textcolor{red}{Results and other stats}

% \href{}{https://docs.google.com/spreadsheets/d/1Ch397EtbmoX15hxgKTI4lUQ2u-ISjL4tznQYEidoLZE/edit?usp=sharing}

\section{Introduction}
\label{sec:introduction}

Social media platforms have enabled people to post and share content online. A significant portion of this content provides valuable resources for initiatives such as citizen journalism, raising public awareness, and supporting political campaigns. However, a considerable amount is posted and shared to mislead social media users and to achieve social, economic, or political agendas. In addition, the freedom to post and share content online has facilitated negative uses, leading to an increase in online hostility, as evidenced by the spread of disinformation, hate speech, propaganda, and cyberbullying~\cite{brooke-2019-condescending, joksimovic-etal-2019-automated, schmidt-wiegand-2017-survey, davidson2017automated, emnlp2019:fine:grained, van-hee-etal-2015-detection}.
A lack of \textit{media literacy}\footnote{Media literacy encompasses the ability to access, analyze, evaluate, and create media in various forms.} is also a major factor contributing to the spread of misleading information on social media \cite{zannu2024influence}. This can lead to the uncritical acceptance and sharing of false or misleading content, which can quickly disseminate through social networks. In their study, \citet{zannu2024influence} highlight the crucial role of media literacy in mitigating the spread of fake news among users of platforms such as Instagram and Twitter. 
% Their research demonstrate the need for enhanced media literacy education to empower users to discern credible information from falsehoods and reduce the propagation of misinformation on social media. 

\begin{figure}[t]
    \centering
    \includegraphics[scale=0.43]{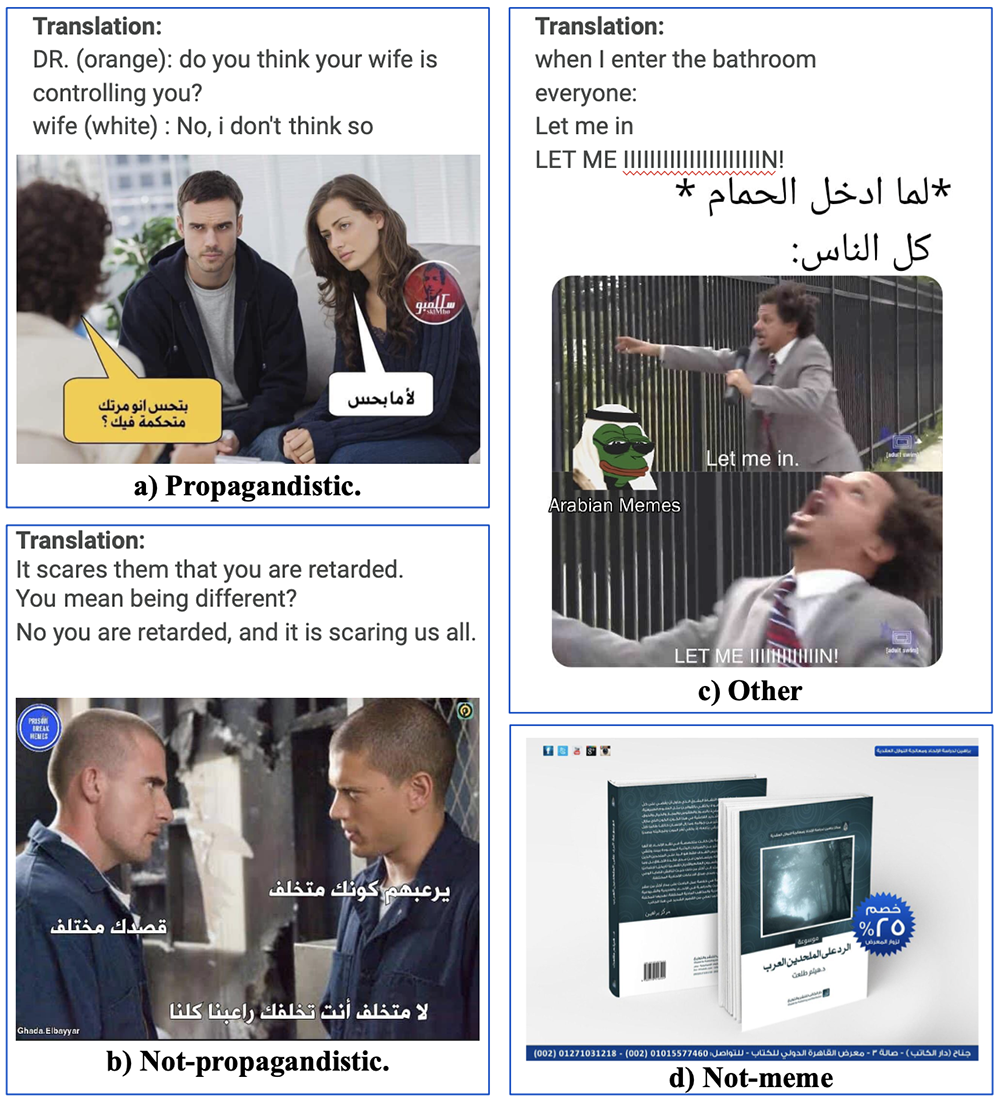}
    \caption{Examples of images representing different categories.}
    \label{fig:example_memes}
\end{figure}

Online content typically consists of different modalities, including text, images, and videos. Disinformation, misinformation, propaganda, and other harmful content are shared across all these modalities. Recently, the use of \textit{Internet memes} have become very popular on these platforms. A meme is defined as ``a collection of digital items that share common characteristics in content, form, or stance, which are created through association and widely circulated, imitated, or transformed over the Internet by numerous users''~\cite{shifman2013memes}. Memes typically consist of one or more images accompanied by textual content~\cite{shifman2013memes, suryawanshi-etal-2020-multimodal}. While memes are primarily intended for humor, they can also convey persuasive narratives or content that may mislead audiences. To automatically identify such content, research efforts have focused on addressing offensive material \cite{Gandhi2020ScalableDO}, identifying hate speech across different modalities \cite{Gomez2020exploring, ChingHatespeechVid2020}, and detecting propaganda techniques in memes \cite{dimitrov2021detecting}. 

Among the various types of misleading and harmful content, the spread of propagandistic content can significantly distort public perception and hinder informed decision-making. 
To address this challenge, research efforts have been specifically directed towards defining techniques and tackling the issue in different types of content, including news articles \cite{da-san-martino-etal-2019-findings}, tweets \cite{propaganda-detection:WANLP2022-overview}, memes \cite{dimitrov2021detecting}, and textual content in multiple languages \cite{piskorski-etal-2023-semeval}. 
Most of these efforts have focused on English, with relatively little attention given to Arabic. Prior research on Arabic textual content includes studies presented at WANLP-2022 and ArabicNLP-2023 \cite{propaganda-detection:WANLP2022-overview,hasanain-etal-2023-araieval}. However, for multimodal content, specifically memes, there are no available datasets or resources. To address this gap, we have collected and annotated a dataset consisting of approximately 6,000 memes, categorizing them into four categories (as shown in Figure \ref{fig:example_memes}) to identify propagandistic content. Below we briefly summarize the contribution of our work. 
\begin{itemize}
    \item The first Arabic meme dataset with manual annotations defining four categories.
    \item A detailed description of the data collection procedure, which can assist the community in future data collection efforts.
    \item An annotation guideline that will serve as a foundation for future research.
    \item Detailed experimental results, including:
    \begin{itemize}
    \item Text modality: training classical models and fine-tuning monolingual vs. multilingual transformer models.
    \item Image modality: fine-tuning CNN models with different architectures.
    \item Multimodality: training an early fusion-based model.
    \item Evaluating different LLMs in a zero-shot setup for all modalities.
    \end{itemize}
    \item Releasing the dataset to the community.\footnote{Dataset is released under CC-BY-NC-SA through \url{https://huggingface.co/datasets/QCRI/ArMeme}.} The dataset and annotation guideline will be beneficial for research to develop automatic systems and enhance media literacy. 
\end{itemize}
% Most of such efforts are focused on English. Compared to that little to no effort has been given for Arabic. Prior research on text modality for Arabic textual content include WANLP-2022 and ArabicNLP-2023 \cite{propaganda-detection:WANLP2022-overview,hasanain-etal-2023-araieval}. For multimodality, specifically for memes, there no dataset and/or reserouces. To fill this gap, in this study, we have collected and annotated a large dataset consisting of $\sim6K$ memes and annotated with four categories to identify propagandistic memes. 

% The associated shared tasks include SemEval-2020 Task 11 on news articles \cite{DaSanMartinoSemeval20task11}, SemEval-2021 Task 6 on memes \cite{SemEval2021-6-Dimitrov}, WANLP-2022 and ArabicNLP-2023 focusing on Arabic \cite{propaganda-detection:WANLP2022-overview,hasanain-etal-2023-araieval}, and SemEval-23 Task 3 on news articles in multiple languages \cite{piskorski-etal-2023-semeval}.

\section{Related Work}
\label{sec:related_work}
% \todo[inline]{@Arid}

Social media has become one of the main ways of sharing information. Its widespread use and reach is also responsible for creating and spreading misinformation and propaganda among users. Propagandistic techniques can be found in various types of content, such as fake news and doctored images, across multiple media platforms, frequently employing tools like bots. Furthermore, such information is distributed in diverse forms, including textual, visual, and multi-modal. To mitigate the impact of propaganda in online media, researchers have been developing resources and tools to identify and debunk such content. 

% The study of propaganda and misinformation can be traced in a wide range of disciplines \cite{jowett2018propaganda} and started in the 17th century to understand the exploitation in various public events \cite{pamphlet_casey, Margolin1979TheVR}. 
% However, over the past few years, researchers have been trying to identify misinformation and propaganda automatically using different techniques from social media content \cite{alam-etal-2022-overview}. 

\subsection{Persuasion Techniques Detection}
Early research on propaganda identification relies on the entire document to identify whether the content is propaganda, while recent studies focus on social media content \cite{SemEval2021-6-Dimitrov}, news articles \cite{emnlp2019:fine:grained}, political speech \cite{partington2017language}, arguments \cite{Habernal.et.al.2017.EMNLP, Habernal2018b}, and multimodal content \cite{dimitrov2021detecting}. \citet{BARRONCEDENO20191849} developed a binary classification (\emph{propaganda} and \emph{non-propaganda}) corpus to explore writing style and readability levels. An alternative approach followed by \citet{Habernal.et.al.2017.EMNLP,Habernal2018b} to identify persuasion techniques within texts constructing a corpus on arguments. Moreover, \citet{emnlp2019:fine:grained} developed a span-level propaganda detection corpus from news articles and annotated in eighteen propaganda techniques. 

\citet{piskorski-etal-2023-multilingual} developed a dataset from online news articles into twenty-two persuasion techniques containing nine languages to address the multilingual research gap. Following the previous work, \citet{piskorski-etal-2023-semeval} and SemEval-2024 task 4 focus on resource development to facilitate the detection of multilingual persuasion techniques.
Focusing on multimodal persuasion techniques for memes, \citet{dimitrov2021detecting} created a corpus containing 950 memes and investigated pretrained models for both unimodal and multimodal memes. The study of \citet{chen2024multimodal} proposed a multimodal visual-textual object graph attention network to detect persuasion techniques from multimodal content using the dataset described in \cite{piskorski-etal-2023-multilingual}. In a recent shared task, \citet{dimitrov2024semeval} introduced a multilingual and multimodal propaganda detection task, which attracted many participants. The participants' systems included various models based on transformers, CNNs, and LLMs.

\subsection{Multimodal Content} 
The study of multimodal content has gained popularity among researchers for propaganda detection due to its effectiveness of in spreading propagandastic information and creating impact among the targeted audience. \citet{ijcai2022p781} presented that propaganda can be used to cause several types of harm including spreading hate, violence, exploitation, etc., while spreading mis- and dis-information is also one of the main reasons \cite{alam-etal-2022-survey}. The study of \citet{Volkova_Ayton_Arendt_Huang_Hutchinson_2019} presented an in-depth analysis of multimodal content for predicting misleading information from news. Additionally, the deception and disinformation analysis on social media platforms using multimodal content in multilingual settings has been studied by \citet{Glenski2019MultilingualMD}. Moreover, hateful memes \cite{kiela2020hateful}, propaganda in visual content \cite{doi:10.1080/15551393.2014.955501}, emotions and propaganda \cite{abd_kadir_etal} also studied by the researchers in the past few years.

Recent studies focusing on fine-tuning visual transformer models such as ViLBERT \cite{lu2019vilbert}, Multimodal Bitransformers \cite{kiela2019supervised}, and VisualBERT \cite{li2019visualbert}. \citet{cao-etal-2022-prompting} study focuses on multimodal hateful meme identification using prompting strategies by adopting \cite{10.1145/3581783.3613836}. \citet{hee2024recent} studied hate speech content moderation and discussed recent advancements leveraging large models.

Compared to previous studies, our work differs in that we provide the first resource for Arabic. Additionally, our annotation guidelines and data collection procedures for memes may be useful for other languages. 

\section{Dataset}
\label{sec:dataset}

\begin{figure*}[]
    \centering
    \includegraphics[scale=0.3]{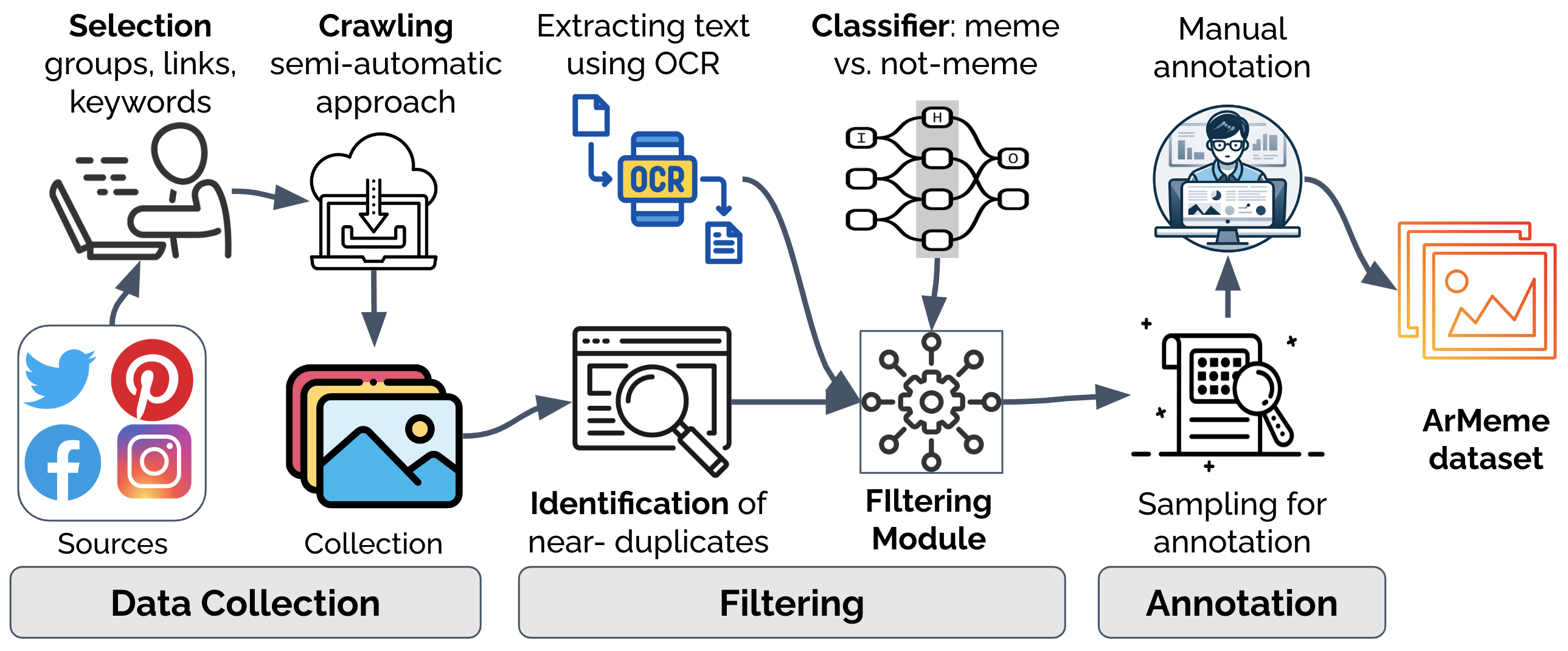}
    \caption{Data curation pipeline.}
    \label{fig:data_collection_pipeline}
\end{figure*}

\subsection{Data Collection}
\begin{figure}[h]
    \centering
    \includegraphics[scale=0.35]{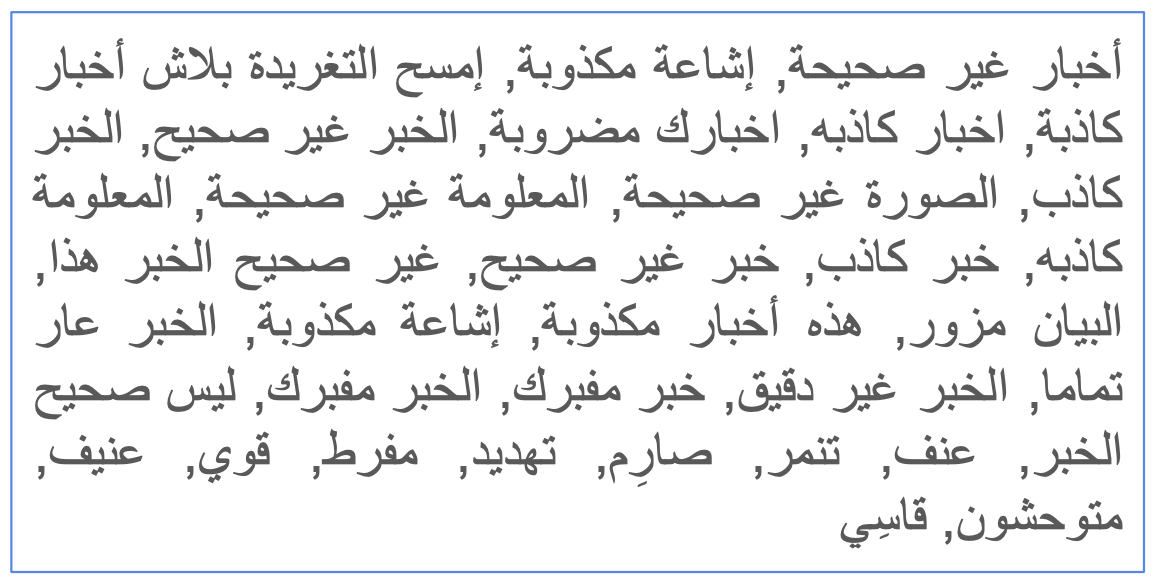}
    \caption{Keywords used to collect tweets.}
    \label{fig:keywords}
\end{figure}

Our data collection process involves several steps as highlighted in the Figure \ref{fig:data_collection_pipeline}. We manually selected public groups and contents from Facebook, Instagram, and Pinterest. In addition, we have also collected memes from Twitter using a set of keywords as listed in the Figure ~\ref{fig:keywords}. Our data curation consists of a series of steps as discussed below. 

% \paragraph{Collection:}  

% \begin{enumerate}
\paragraph{\textbf{Manual selection of groups, links, and keywords:}} Focusing on the mentioned sources, we have manually selected public groups, which contains post on public figures, celebrity, and discussions about politics. In Table \ref{tab:data_collection_stat}, we provide the sources of the dataset, number of groups, and number of images we have collected. 
\begin{table}[h]
\centering
\begin{tabular}{@{}lrr@{}}
\toprule
\multicolumn{1}{c}{\textbf{Source}} & \multicolumn{1}{c}{\textbf{\# of Group}} & \multicolumn{1}{c}{\textbf{\# of Images}} \\ \midrule
Facebook & 19 & 5,453 \\
Instagram & 22 & 107,307 \\
Pinterest & - & 11,369 \\
Twitter & - & 5,369 \\ \midrule
\textbf{Total} &  & \textbf{129,498} \\ \bottomrule
\end{tabular}
\caption{Statistics of the initial data collection.}
\label{tab:data_collection_stat}
\end{table}

\paragraph{\textbf{Crawling:}} Given that Facebook, Instagram, and Pinterest do not provide API or do not allow automatic crawling images, therefore, we developed a semi-automatic approach to crawl images from these platforms. The steps include manually loading images and then crawl the images that are loaded on the browser. For the Twitter (X-platform), we used the keywords to crawl tweets, which consists of media/image.

\subsection{Filtering}
\paragraph{\textbf{Filtering duplicate images:}} Given that user might have posted same meme or a slight modification of it in multiple platforms, which is very common for social media, therefore, we applied an exact and near-duplicate image detection method to remove them. This method consists of extracting features using a pre-trained deep learning model and compute similarity. Given a dataset $\mathcal{D} = \{x_1, x_2, \ldots, x_N\}$ consisting of $N$ data points, we extracted features using a pre-trained deep learning model and used nearest neighbor based approach~\cite{cunningham2007k}. The model is trained by fine-tuning ResNet18~\cite{he2016deep} using the social media dataset discussed in \cite{alam2020deep}. Let $f: \mathbb{R}^d \rightarrow \mathbb{R}^m$ be a pre-trained deep learning model that maps an input data point $x_i \in \mathbb{R}^d$ to a feature vector $f(x_i) \in \mathbb{R}^m$. For each data point $x_i \in \mathcal{D}$, the feature vector is extracted as: $\mathbf{z}_i = f(x_i), \quad \text{for } i = 1, 2, \ldots, N$
where $\mathbf{z}_i \in \mathbb{R}^m$ is the feature vector of the data point $x_i$. To compute the nearest neighbors between a data point $x_i$ and the entire dataset $\mathcal{D}$, we use the euclidean distance. We then use a threshold of $3.6$ to define the near-duplicate images as those with a euclidean distance less than or equal to this threshold value.

\paragraph{\textbf{OCR Text:}} We used EasyOCR\footnote{\url{https://github.com/JaidedAI/EasyOCR}} to extract text from memes. Memes with no extracted or detected text were filtered out. The reasons for choosing EasyOCR are: (a) it is a ready-to-use OCR with 80+ supported languages and all popular writing scripts, and (b) it includes the implementation of the state-of-the-art, highly efficient real-time scene text detection module \cite{liao2022real}, called DBnet, which uses differentiable binarization and adaptive scale fusion.
    
\paragraph{\textbf{Classifier-Based Filtering:}} 
% \textcolor{blue}{
We employed an in-house meme vs. non-meme classifier to filter out images that were not classified as memes. The classifier was developed using a dataset of 3,935 images, consisting of 2,000 memes and 1,935 non-memes. Following the approach of \cite{hasnat2019application}, we developed a lightweight meme classifier to perform binary classification based on the extracted image features. The classifier achieved the best performance of 94.79\% test set accuracy in classifying memes using a 256-dimensional normalized histogram extracted from gray-scale images as features, with a Multilayer Perceptron (MLP) as the classifier.
% }

\subsection{Annotation}

\paragraph{\textbf{Data Sampling:}} Due to budget constraints for manual annotation, we randomly sampled $\sim$6K images.

\paragraph{\textbf{Manual Annotation:}} 
For the manual annotation, we first prepared an annotation guideline to assist the annotators. To facilitate the annotation tasks, we developed an annotation platform as presented in Appendix \ref{sec:app_annotation_platform}. The details of the annotation guidelines are reported in Appendix \ref{sec:annotation_instruct_english}. 
Note that we developed the annotation guidelines in English, (see Section \ref{sec:annotation_instruct_english}), which were then translated into the Arabic language. Translating the guideline in native language was indeed important and also inspired by prior work \cite{alam2021fighting,hasanain2024can}. The idea is not only to make the annotation task more convenient but also capture different linguistic aspects. 
The guidelines included several examples of memes. It was reviewed by several NLP experts who are also native Arabic speakers. 
% The details of the Arabic annotation guideline can be found in \url{https://shorturl.at/3z4CS}.
% https://docs.google.com/document/d/1vY_hK6FF1h4sMTEaPGfk6m3tJ3rrLL9f/edit

In Figure \ref{fig:example_memes}, we provide examples of images and memes representing different categories. Figure \ref{fig:example_memes}(a) depicts a couple in what appears to be a couples therapy session. The therapist asks the husband, ``Do you feel your wife is controlling you?'' The wife responds, ``No, I don’t feel so.'' It is evident that the question was directed towards the husband, yet the wife answers instead of him. The irony lies in her controlling the conversation when her control is the subject of discussion. This meme attempts to humorously portray the stereotypical notion that wives are controlling in marriage. Figure \ref{fig:example_memes}(b) employs a play on words to create humor. The Arabic word that means "different" is similar to the Arabic word "retarded", except in the position of two letters. However, this meme does not contain any propagandistic techniques. Figure \ref{fig:example_memes}(c) features a meme that uses an image of a scene from TV with dialogues and added text to create humor. However, it was categorized as ``other'' because the dialogues were in English, rather than ``not propagandistic'' or ``propagandistic''. Figure \ref{fig:example_memes}(d) shows a picture of book covers, which might have been part of an advertisement. This image was labelled as "not meme". 

The annotation tasks consist of two phases: 
\begin{itemize}
    \item Phase 1 (image categorization): labeling images shown on the platform as \textit{(i)} not-meme, \textit{(ii)} other, \textit{(iii)} not propaganda, or \textit{(iv)} propaganda. Each image was annotated by three annotators and the final label is decided based on majority agreement.
    \item Phase 2 (text editing): editing the text to fix OCR errors. This step was performed only for images labelled as meme, and propagandistic or not propagandistic. Further details is mentioned in \ref{ssec:annotation_meme_categorization}.  
\end{itemize}

% \textbf{(1)  
% \textbf{(2)  
% \footnote{} %provided in Section (Appendix Section E). 

\noindent\textbf{Annotation Team:} 
% We recruited local annotators, who 
% \todo[inline]{@Fatema, please revise..}
The team in phase 1 consisted of three members, and in phase 2, consisted of one member. All annotators are native Arabic speakers holding at least a bachelor's degree. Our in-house expert annotator provided them with several iterations of training, supervised and monitored their work, and handled quality control throughout the entire annotation process. This quality assurance included periodic checks of random annotation samples and providing feedback. 
Since the institute requires the signing of a Non-Disclosure Agreement (NDA), each annotator signed an NDA after being made aware of the institute's terms and conditions. They were compensated at the same rate as charged by external companies. 

% \todo[]{Payment related to annotation}
% \todo[]{demographics}
% \todo[]{NDA sign}

\noindent\textbf{Annotation platform:} We utilized our in-house annotation platform for the annotation task. Separate annotation interfaces were designed for each phase.

% \todo[inline]{Fatema, please explain the figure \ref{fig:example_memes}}

\paragraph{Annotation Agreement}
For the Phase 1 annotation, we computed annotation agreement using various evaluation measures, including Fleiss' kappa, Krippendorff’s alpha, average observed agreement, and majority agreement. The resulting scores were 0.529, 0.528, 0.755, and 0.873, respectively. Based on the value of Krippendorff’s alpha, we can conclude that our annotation agreement score indicates moderate agreement.\footnote{Note that Kappa values of 0.21--0.40, 0.41--0.60, 0.61--0.80, and 0.81--1.0 correspond to fair, moderate, substantial, and perfect agreement, respectively~\cite{landis1977measurement}.} In the final label selection, we excluded the $\sim$200 memes on which the annotators disagreed. In the \textit{second phase}, we mainly edited text to fix the OCR errors, which has been done by a single annotator. To ensure the quality of the \textit{editing phase}, random samples were checked by an expert annotator and periodically provided feedback. Note that the post-editing has been done for only propagandistic and non-propagandistic memes. It is to reduce the cost of the annotation, and to further annotate them with span-level propaganda techniques. 

% \begin{enumerate} %[leftmargin=*]
% \itemsep-0.1em 
%     \item Fleiss kappa: It is a reliability measure that is applicable for any fixed number of annotators annotating categorical labels to a fixed number of items, which can handle two or more categories and annotators~\cite{fleiss2013statistical}. However, it can not handle missing labels, except for excluding them from the computation. 
%     \item Average observed agreement: It is an average observed agreement over all pairs of annotators \cite{fleiss2013statistical}.
%     \item Majority agreement: We compute the majority at the tweet level and take the average. The reason behind this is that for many tweets the number of annotators vary between three and five, and hence, it is plausible to evaluate the agreement at the tweet level.
%     \item Krippendorff’s alpha: It is a measure of agreement that allows two or more annotators and categories~\cite{krippendorff1970estimating}. Additionally, it handles missing labels. 
% \end{enumerate}

\subsection{Statistics} 
Table \ref{tab:data_split_stat} shows the number of memes for each category. For the rest of the experiments, the data was split into train, dev, and test as shown in the table. The dataset comprises a total of 5,725 annotated samples, with ``Not propaganda'' covers over half of the dataset ($\sim$66\%), followed by ``Propaganda.'' The ``Not-meme`` and ``Other`` classes are significantly smaller in comparison. The distribution indicates a significant class imbalance, particularly between ``Not propaganda'' and the other classes, which could affect model training and performance.

In Table \ref{tab:source_wise_dist}, we report the distribution of the dataset across different sources. The annotated number of memes reflects the memes we collected from various sources, as detailed in Table \ref{tab:data_collection_stat}. We have the highest number of memes collected and annotated from Instagram. A very small number from Twitter is due to different image filtering steps. As shown in Table \ref{tab:source_wise_dist} the prevalence of propagandistic memes is relatively higher on Facebook than that of non-propagandistic memes.

\begin{table}[]
\centering
\scalebox{0.95}{%
\begin{tabular}{@{}lrrrr@{}}
\toprule
\multicolumn{1}{c}{\textbf{Class label}} & \multicolumn{1}{c}{\textbf{Train}} & \multicolumn{1}{c}{\textbf{Dev}} & \multicolumn{1}{c}{\textbf{Test}} & \multicolumn{1}{c}{\textbf{Total}} \\ \midrule
Not propaganda & 2,634 & 384 & 746 & 3,764 \\
Propaganda & 972 & 141 & 275 & 1,388 \\
Not-meme & 199 & 30 & 57 & 286 \\
Other & 202 & 29 & 56 & 287 \\ \midrule
\textbf{Total} & \textbf{4,007} & \textbf{584} & \textbf{1,134} & \textbf{5,725} \\ \bottomrule
\end{tabular}
}
\caption{Data split statistics.}
\label{tab:data_split_stat}
\end{table}

\begin{table}[]
\centering
\setlength{\tabcolsep}{2pt}
\scalebox{0.9}{%
\begin{tabular}{@{}lrrrrr@{}}
\toprule
\multicolumn{1}{c}{\textbf{Source}} & \multicolumn{1}{c}{\textbf{Not prop.}} & \multicolumn{1}{c}{\textbf{Prop.}} & \multicolumn{1}{c}{\textbf{Not-meme}} & \multicolumn{1}{c}{\textbf{Other}} & \multicolumn{1}{c}{\textbf{Total}} \\ \midrule
Facebook & 464 & 332 & 58 & 144 & 998 \\
Instagram & 2,052 & 637 & 46 & 60 & 2,795 \\
Pinterest & 1,245 & 414 & 147 & 78 & 1,884 \\
Twitter & 3 & 5 & 38 & 2 & 48 \\ \midrule
\textbf{Total} & \textbf{3,764} & \textbf{1,388} & \textbf{289} & \textbf{284} & \textbf{5,725} \\ \bottomrule
\end{tabular}
}
\caption{Number of annotated memes across different sources. Prop. - Propaganda.}
\label{tab:source_wise_dist}
\end{table}
\section{Experiments}

\subsection{Training and Evaluation Setup}
For all experiments, except for those involving LLMs as detailed below, we trained the models using the training set, fine-tuned the parameters with the development set, and assessed their performance on the test set. We use the model with the best weighted-F1 on the development set to evaluate its performance on the test set. For the LLMs, we accessed them through APIs.

\paragraph{Evaluation Measures}
For the performance measure for all different experimental settings, we compute accuracy, and weighted precision, recall and F$_1$ score. In addition, we also computed macro-F1. 

\subsection{Models}
We conducted our experiments using classical models (e.g., SVM) as well as both small (e.g., ConvNeXt-T) and large language models. It is important to note that our definitions of `small' and `large' models are based on the criteria discussed in \cite{zhao2023survey}.\footnote{The term `LLMs' specifically refers to models that encompass tens or hundreds of billions of parameters.} 

% \subsection{Models}
% \todo[inline]{@Arid}

\subsubsection{Baseline:} We adopted widely-used standard baseline methods, including the majority and random baselines.
 
\subsubsection{Small Language Models (SLMs)} 
We implemented classical models across all modalities, consisting of \emph{(i)} feature extraction followed by model training, and \emph{(ii)} fine-tuning pre-trained models (PLMs). For fine-tuning PLMs, we used a task-specific classification head over the training subset.

\paragraph{Text-Based Models:} 
 For the text-based unimodal model, we transformed text into $n$-gram (n=1) format using a tf-idf representation, considering the top 5,000 tokens, and trained an SVM model with a parameter value of \( C = 1 \). Additionally, we fine-tuned several pre-trained transformer models (PLMs). These included the monolingual transformer model AraBERT \citep{baly2020arabert}, Qarib \cite{abdelali2021pretraining} and multilingual transformers such as multilingual BERT (mBERT)~\cite{devlin2018bert}, and XLM-RoBERTa (XLM-r)~\citep{conneau2019unsupervised}. We used the Transformer toolkit~\cite{Wolf2019HuggingFacesTS} for the experiment. 
 Following the guidelines outlined in \citep{devlin2018bert}, we fine-tuned each model using the default settings over three epochs. Due to instability, we performed ten reruns for each experiment using different random seeds, and we picked the model that performed best on the development set. We provided the details of the parameters settings in Appendix \ref{sec:details_of_exp}.

\paragraph{Image-Based Models:}
For the image-based unimodal model with feature-extraction approach, we extracted features using ConvNeXt-T \cite{liu2022convnet},\footnote{The configuration of ConvNeXt-T includes \( C = (96, 192, 384, 768) \) and \( B = (3, 3, 9, 3) \), where \( C \) and \( B \) represent the number of channels and blocks, respectively.} and trained an SVM model. For fine-tuning image-based PLMs, we used ResNet18, ResNet50~\cite{he2016deep}, VGG16~\cite{simonyan2014very}, MobileNet~\cite{howard2017mobilenets}, and EfficientNet~\cite{tan2019efficientnet}. We chose these diverse architectures to understand their relative performance. The models were trained using the Adam optimizer~\cite{kingma2014adam} with an initial learning rate of $10^{-3}$, which was decreased by a factor of 10 when accuracy on the development set stopped improving for 10 epochs. The training lasted for 150 epochs.

% DenseNet~\cite{huang2017densely}, SqueezeNet~\cite{i2016squeezenet}, 

\paragraph{Multimodal Models:}
We developed a multimodal model by concatenating text features (extracted using AraBERT) and image features (extracted using ConvNeXt-T), which were then fed into an SVM.

% \cite{chung2022scaling}
% \cite{kowsher2022bangla}

\subsubsection{LLMs for Text}
For the LLMs, we investigate their performance with 
% in-context 
zero-shot 
learning settings without any specific training. It involves prompting and post-processing of output to extract the expected content. Therefore, for each task, we experimented with a number of prompts.
% , guided by the same instruction and format as recommended in the OpenAI Chat playground, and PromptSource~\cite{bach2022promptsource}. 
We used 
% the following models: Flan-T5 (large and XL) \cite{chung2022scaling}, BLOOMZ (1.7B, 3B, 7.1B, 176B-8bit) \cite{muennighoff2022crosslingual} 
GPT-4~\cite{openai2023gpt}. We set the temperatures to zero for all these models to ensure deterministic predictions. We used LLMeBench framework \cite{dalvi-etal-2024-llmebench} for the experiments, which provides seamless access to the API end-points and followed prompting approach reported in \cite{abdelali-etal-2024-larabench}.

\subsubsection{Multimodal LLMs}
% and Vision-Language Models (VLMs)}
% \textcolor{blue}{
For the multimodal models \cite{xu2023multimodal}, we experimented with several well-known and top-performing commercial models. These included OpenAI's GPT models (GPT-4 Turbo and GPT-4o) \cite{openai2023gpt}, as well as Google's Gemini Pro models (versions 1.0 and 1.5) \cite{team2023gemini}.

% For the VLMs \cite{zhang2024vision}, we experimented with several state-of-the-art models, such as CogVML \cite{wang2023cogvlm} and BakLLaVA \cite{liu2024visual}. 
Using these models, we tested \emph{(i)} the meme/image only, \emph{(ii)} text only (text extracted using OCR from the image), and \emph{(iii)} multimodal (meme and OCR text) in a zero-shot learning setting. This means we did not provide any training examples within the prompts to the models.
% }
%GPT-4o  model version: 2024-05-13
% vision-preview

%
% \textcolor{blue}{
We designed a prompt based on trial and error using the visual interfaces of OpenAI's GPT-4 user interface. The prompt instructs the models to perform a deeper analysis of the image and any text that they can read within the image before answering whether the meme can be classified as spreading propaganda. Additionally, it requests the models to provide the output in a valid JSON format. For the experiments, we used the default parameters for each multimodal model. 
% and VLM. 

% The input to these models is \emph{(i)} the image only, \emph{(ii)} text only, and \emph{(iii)} multimodal, as the models can extract the text and include it while analyzing the content and providing the answer. 
% Note that, we frequently experienced difficulties obtaining outputs from the models (except GPT-4o). 

% Therefore, we applied 10 retries for each sample until we obtained a valid output. Despite this, BakLLaVA \cite{liu2024visual} did not provide valid outputs from which we could extract the classification results, and hence we did not include this model in the competitive results. 

%
%
% }

% \paragraph{Zero-shot Learning} We used GPT-4~\cite{openai2023gpt} for zero-shot learning which is also a transformer-based model. We simply used the test set for evaluating the GPT-4 model without using any kind of training data. As for the prompt, we used the similar format discussed in \cite{abdelali2023benchmarking}, as also shown in Listing \ref{lst:example}. Our prompt was relatively simple, which can be explored further in future studies. 

\subsection{Prompting Strategy}
LLMs produce varied responses depending on the prompt design, which is a complex and iterative process that presents challenges due to the unknown representation of information within different LLMs. The instructions expressed in our prompts include English language with the input text content in Arabic. 

% In Appendix \ref{sec:appx-prompts} we provide examples of prompts for different tasks. 
% \subsubsection{Zero-shot}
As mentioned earlier we employed zero-shot prompting, providing natural language instructions that describe the task and specify the expected output. This approach enables the LLMs to construct a context that refines the inference space, yielding a more accurate output. In Listing \ref{lst:zero_shot_gpt4}, we provide an example of a zero-shot prompt, emphasizing the instructions and placeholders for both input and label. 
% Given that GPT-4 has the capability to play a role, therefore, we also provide a role for it as an ``expert social media analyzer''
Along with the instruction we provide the labels to guide the LLMs and provide information on how the LLMs should present their output, aiming to eliminate the need for post-processing.

% In our initial set of experiments with BLOOMZ, we observed that it did not respond as effectively to the same instructions as GPT-4. Therefore, we used more straightforward instructions for BLOOMZ, as illustrated in Listing \ref{lst:zero_shot_Bloomz}. For the other versions of BLOOMZ and Flan-T5, we used the same prompt as BLOOMZ. 

\begin{lstlisting}[caption={Zero-shot prompt example for GPT-4.}, label=lst:zero_shot_gpt4]
Instructions:
prompt = (
"You are an expert social media image analyzer specializing in identifying propaganda in Arabic contexts. "
"I will provide you with Arabic memes and the text extracted from these images. Your task is to briefly analyze them. "
"To accurately perform this task, you will: (a) Explicitly focus on the image content to understand the context and provide a meaningful description and "
"(b) pay close attention to the extracted text to enrich your description and support your analysis. "
"Finally, provide response in valid JSON format with two fields with a format: {\"description\": \"text\", \"classification\": \"propaganda\"}. Output only json. "
"The \"description\" should be very short in maximum 100 words and \"classification\" label should be \"propaganda\" or \"not-propaganda\" or \"not-meme\" or \"other\". "
"Note, other is a category, which is used to label the image that does not fall in any of the previous category."
)  

\end{lstlisting}

\begin{table}[t]
\centering
\setlength{\tabcolsep}{2pt}
\scalebox{0.84}{%
\begin{tabular}{@{}lrrrrr@{}}
\toprule
\multicolumn{1}{c}{\textbf{Model}} & \multicolumn{1}{c}{\textbf{Acc}} & \multicolumn{1}{c}{\textbf{W-P}} & \multicolumn{1}{c}{\textbf{W-R}} & \multicolumn{1}{c}{\textbf{W-F1}} & \multicolumn{1}{c}{\textbf{M-F1}} \\ \midrule
\multicolumn{6}{c}{\textbf{Baseline}} \\  \midrule
Majority & 0.658 & 0.433 & 0.659 & 0.522 & 0.198 \\
Random & 0.479 & 0.518 & 0.479 & 0.479 & 0.239 \\  \midrule
\multicolumn{6}{c}{\textbf{Unimodal - Text}} \\  \midrule
Ngram & 0.669 & 0.624 & 0.669 & 0.582 & 0.280 \\
AraBERT & 0.688 & 0.670 & 0.688 & 0.666 & 0.511 \\
Qarib & 0.697 & 0.688 & 0.697 & \textbf{0.690} & \textbf{0.551} \\
mBERT & 0.707 & 0.688 & 0.707 & 0.675 & 0.487 \\
XLM-r & 0.699 & 0.676 & 0.699 & 0.678 & 0.489 \\
GPT-4v & 0.664 & 0.620 & 0.664 & 0.624 & 0.384 \\ 
GPT-4o & 0.573 & 0.611 & 0.573 & 0.579 & 0.350 \\ 
\midrule
\multicolumn{6}{c}{\textbf{Unimodal - Image}} \\ \midrule
CNeXt + SVM & 0.655 & 0.608 & 0.655 & 0.614 & 0.405 \\
MobileNet (v2) & 0.660 & 0.618 & 0.660 & 0.620 & 0.426 \\
ResNet18 & 0.656 & 0.597 & 0.656 & 0.593 & 0.358 \\
ResNet50 & 0.660 & 0.638 & 0.660 & 0.637 & 0.434 \\
Vgg16 & 0.656 & 0.597 & 0.656 & 0.593 & 0.358 \\
Eff (b7) & 0.660 & 0.597 & 0.660 & 0.595 & 0.352 \\ 
GPT-4v & 0.565 & 0.551 & 0.565 & 0.545 & 0.223 \\ 
GPT-4o & 0.693 & 0.627 & 0.693 & 0.634 & 0.305 \\ \midrule
\multicolumn{6}{c}{\textbf{Multimodal}} \\ \midrule
CNeXt + ArB + SVM & 0.683 & 0.655 & 0.683 & 0.659 & 0.513 \\
Gemini & 0.519 & 0.551 & 0.519 & 0.521 & 0.276 \\ 
GPT-4v & 0.681 & 0.461 & 0.330 & 0.619 & 0.340 \\
GPT-4o & 0.653 & 0.443 & 0.354 & 0.639 & 0.363 \\
\bottomrule
\end{tabular}
}
\caption{Classification with different modalities. CNeXt: ConvNeXt, Eff (b7): Efficientnet (b7), Gemini: Gemini-1.5-flash-preview-0514l, GPT-4v: GPT-4-vision (gpt-4-vision-preview) W-*: weighted average; M-: Macro average. XLM-r: XLM-RoBERTa base.}
\label{tab:classification_results}
\end{table}

\section{Results and Discussion}
\label{label:results}
In Table \ref{tab:classification_results}, we report the detailed classification results for different modalities and models. All models outperform the majority and random baselines. Among the text-based models, the fine-tuned Qarib model outperforms all other models, achieving the best results (\textbf{0.690} weighted F1) across all modalities and models. AraBERT is the second-best fine-tuned model, with a weighted F1-score of 0.666 among the text-based models. The performance of multilingual transformer models is relatively worse than that of monolingual models.

For the image-based models, the fine-tuned ResNet50 shows the best result (\textbf{0.673} weighted F1) among all other fine-tuned models and GPT-4o model. The performance of MobileNet (v2) and \textbf{CNeXt + SVM} rank as the second and third best among the fine-tuned models. The results of VGG16 and EfficientNet (b7) are almost similar.

For the multimodal models, the model trained with \textit{ConvNeXt + AraBERT + SVM} shows the highest performance (0.659 weighted F1) among the multimodal LLMs. The performance of Gemini is significantly worse than that of the GPT-4 variants. GPT-4o demonstrates higher performance compared to GPT-4 Vision.

In our experiments all multimodal model are tested using zero-shot setting, therefore, such lower performance compared to the fine-tuned models are expected.

\section{Additional Experiments}
We further conducted experiments using the dataset released as part of the ArAIEval shared task 2 \cite{araieval:arabicnlp2024-overview}, focusing on two labels: propaganda and not-propaganda. The dataset statistics are provided in Table \ref{tab:dataset_task2}. The goal was to investigate model performance in a binary classification scenario and we benchmarked this dataset using multimodal models.

% \subsection{ArAIEval Shared Task Dataset}
% \label{app:dataset_araieval}
% Table \ref{tab:dataset_task2} report the distribution of the dataset released as a part of the ArAIEval shared task. 
% For our study 
\begin{table}[h]
\centering
\begin{tabular}{@{}lrrrr@{}}
\toprule
\multicolumn{1}{c}{\textbf{Class labels}} & \multicolumn{1}{c}{\textbf{Train}} & \multicolumn{1}{c}{\textbf{Dev}} & \multicolumn{1}{c}{\textbf{Test}} & \multicolumn{1}{c}{\textbf{Total}} \\ \midrule
Not propaganda & 1,540 & 224 & 436 & 2,200 \\
Propaganda & 603 & 88 & 171 & 862 \\
\textbf{Total} & 2,143 & 312 & 607 & 3,062 \\ \bottomrule
\end{tabular}
\caption{Distribution of dataset for ArAIEval shared task 2.}
\label{tab:dataset_task2}
\end{table}

% In order to address this issue for the competitive models, we included a new metric, \textit{Num. Failures}, along with the other evaluation metrics in Table \ref{tab:res_vlms}.

% \subsubsection{Multimodal and Vision-Language Models (VLMs)}
% \label{sec:results}

% \textcolor{blue}{
 %. and VLMs 
Table \ref{tab:results_araieval_dataset} presents the competitive results of three multimodal models with image-only input: GPT-4o, GPT-4 Turbo, and Gemini Pro 1.0. Among these models, GPT-4o significantly outperforms the others and demonstrates the highest performance across all evaluated metrics, achieving an accuracy of 85.17\%, a precision of 84.80, a recall of 85.17, and a weighted F1-score of 84.87. In comparison, GPT-4 Turbo lags behind GPT-4o in all metrics, with an accuracy of 76.44\%, indicating a significant performance drop compared to GPT-4o. Gemini Pro 1.0 shows lower performance than the GPT-4 models, with an accuracy of 72.47\%.

% and 33 failures. 
% Finally, CogVML shows slightly better performance than Gemini Pro 1.0 in accuracy (73.09\%) but suffers from a significantly higher number of failures, totaling 161.

%also shows no failures but 
% }

% \textcolor{blue}{
% The number of failures highlights an important aspect of model reliability and robustness. Despite their lower performance compared to GPT-4o, Gemini Pro 1.0 and CogVML exhibit substantial issues with reliability, as evidenced by their higher failure rates, with CogVML being particularly problematic.
% }

\begin{table}[h]
\centering
\setlength{\tabcolsep}{3pt}
\scalebox{1.0}{
\begin{tabular}{@{}lrrrrr@{}}
\toprule
\multicolumn{1}{c}{\textbf{Model}} & \multicolumn{1}{c}{\textbf{Acc.}} & \multicolumn{1}{c}{\textbf{W-P}} & \multicolumn{1}{c}{\textbf{W-R}} & \multicolumn{1}{c}{\textbf{W-F1}} & \multicolumn{1}{c}{\textbf{M-F1}}\\ \midrule
Gemini & 0.725 & 0.685 & 0.725 & 0.663 & 0.345\\ 
GPT-4v & 0.764 & 0.748 & 0.764 & 0.735 & 0.645\\
GPT-4o & 0.852 & 0.848 & 0.852 & \textbf{0.849} & 0.810\\
\bottomrule
\end{tabular}
}
\caption{Results on ArAIEval dataset. Gemini: version Pro 1.0.}
\label{tab:results_araieval_dataset}
\end{table}
\section{Conclusions and Future Work}
\label{sec:conclusions}
In this study, we introduce a manually annotated dataset for detecting propaganda in Arabic memes. We have annotated $\sim$ 6K memes with four different categories, making it the first such resource for Arabic content. To facilitate future annotation efforts for this type of content, we developed annotation guidelines in both English and Arabic and are releasing them to the community. Our work provides an in-depth analysis of the dataset and includes extensive experiments focusing on different modalities and models, including pre-trained language models (PLMs), large language models (LLMs), and multimodal LLMs. Our results indicate that fine-tuned models significantly outperform LLMs.

In future work, we plan to extend the dataset with further annotations that include hateful, offensive, and propagandistic techniques.

\section{Limitations}
The dataset we have collected originates from various public groups on Facebook, Instagram, Pinterest, and Twitter. The annotated dataset is highly imbalanced, which may affect model performance. Therefore, it is important to develop models with this aspect in mind.

\section*{Ethics and Broader Impact}
Our dataset solely comprises memes, and we have not collected any user information; therefore, the privacy risk is nonexistent. It is important to note that annotations are subjective, which inevitably introduces biases into our dataset. However, our clear annotation schema and instructions aim to minimize these biases. We urge researchers and users of this dataset to remain critical of its potential limitations when developing models or conducting further research. Models developed using this dataset could be invaluable to fact-checkers, journalists, and social media platforms.

\section*{Acknowledgments}
The work of F. Alam, M. Hasanain, and F. Ahmed is supported by the NPRP grant 14C-0916-210015 from the Qatar National Research Fund part of Qatar Research Development and Innovation Council (QRDI). The findings achieved herein are solely the responsibility of the authors.

% \bibliographystyle{acl_natbib}
\bibliography{bib/main,bib/anthology} 

% \clearpage
% \newpage
\appendix

\section{Details of The Experiments}
\label{sec:details_of_exp}
For the experiments with transformer models, we adhered to the following hyper-parameters during the fine-tuning process. Additionally, we have released all our scripts for the reproducibility.

\begin{itemize}[nosep]
    \item Batch size: 8;
    \item Learning rate (Adam): 2e-5;
    \item Number of epochs: 10;
    \item Max seq length: 256.
\end{itemize}

\textbf{Models and Parameters:}
\begin{itemize}[nosep]
    \item \textbf{AraBERT}: L=12, H=768, A=12; the total number of parameters is 371M.
    \item \textbf{XLM-RoBERTa} (xlm-roberta-base): L=24, H=1027, A=16; the total number of parameters is 355M.
    % \item \textbf{BLOOMZ} (bigscience/bloom-560m): L=24, H=1024, A=16; the total number of parameters is 560M.
    % \item \textbf{BLOOMZ} (bigscience/bloom-1b7): L=24, H=2048, A=16; the total number of parameters is 1.7B.
\end{itemize}

\section{Annotation Task} % in English to Categorize Memes}
\label{sec:annotation_instruct_english}
% \todo[inline]{@Fatema, please write a complete details of the annotation steps both phase 1 and 2 with annotation interface, need good examples in the screen-shots, (1) One example of categorization and (2) text editing: before and after}
We designed the annotation instructions through careful analysis and discussion, followed by iterative refinements based on observations and input from the annotators based on the pilot annotation. Our annotation schema is structured into two phases as discussed below. 
% as shown in Figure \ref{fig:annotation_steps}.

\subsection{Phases of Annotations}
\begin{figure}[]
    \centering
    \includegraphics[scale=0.3]{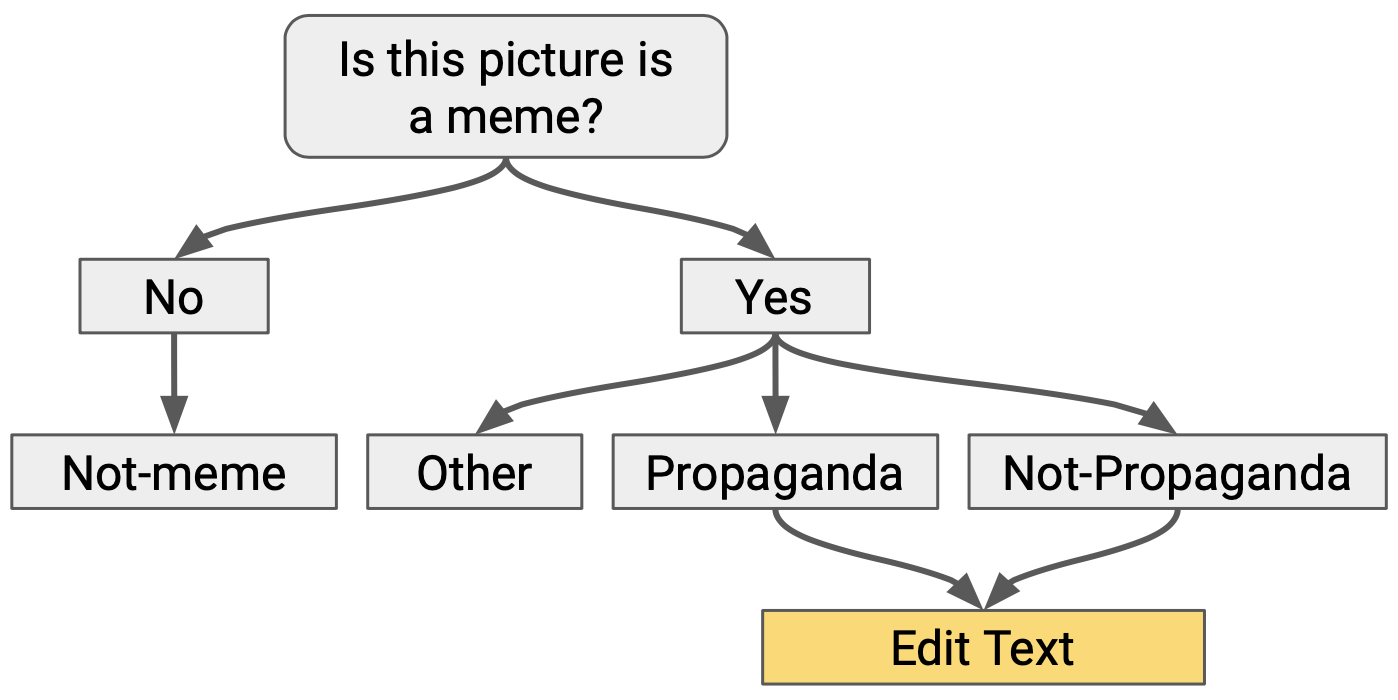}
    \caption{A visual representation of the annotation process. Block with yellow color represents phase 2.}
    \label{fig:annotation_steps}
\end{figure}

To ensure the quality of the annotation and facilitate the work of annotators, we conducted the annotation in two phases: \textit{(i)} image categorization and \textit{(ii)} text editing. The \textit{first phase} (see Section \ref{ssec:annotation_meme_categorization}) focuses primarily on categorizing the images shown on the interface. In the second phase (see Section \ref{ssec:app_instructions_text_editing}), our goal is to edit the text that can be seen on the images only for images that were labelled as memes and as propagandistic or not propagandistic. The motivation for editing the text for these categories is to further utilize them for other annotation tasks. For example, propagandistic memes can be further annotated with specific propagandistic techniques. In Figure \ref{fig:annotation_steps}, we illustrate the thought process of the meme annotation phases.

% The mental working processing of the whole annotation task is outlined in Figure \ref{fig:annotation_steps}.

\subsection{Meme Categorization}
\label{ssec:annotation_meme_categorization}

\subsubsection{Definition of a Meme:}
% This task will rely on the following definition of a meme:
Memes typically consist of a background image, which could be a photograph, illustration, or screenshot, and a layer of text that adds context, humor, or commentary to the image. The text is usually placed at the top and/or bottom of the image but not always. The combination of the image and the text creates a specific message, joke, or commentary that is meant to be easily understood, relatable, and shareable.
% \textbf{How to identify if the picture shown is a meme? }
Some characteristics of memes as observed during analysis and discussion: 
% \begin{enumerate} %[noitemsep,topsep=0pt,leftmargin=*,labelwidth=!,labelsep=.5em]
%     \item Text overlaid on image.
%     \item The text has humor in it. 
%     \item The picture \textit{must} meet points 1 and 2.
% \end{enumerate}
% \textbf{Some characteristics of memes:}

\begin{enumerate} %[noitemsep,topsep=0pt,leftmargin=*,labelwidth=!,labelsep=.5em]
    \item contains text overlaid on image.
    \item The text has humor in it. 
    \item The image \textit{must} meet points 1 and 2.
    \item Some contents of the image have been edited.
    \item Text might be added to different locations of the image. 
    \item May use images of entities with facial expressions (human, animals, fictional characters,etc.), which are then used to construct meaning alongside the added text. 
    \item May use an entity performing a certain action that might be used to construct meaning alongside the added text.
    \item May use an entity that represents an idea or culture, to construct meaning alongside the added text.
    \item May use screenshots from movie scenes and dialogues with added comments to create memes. 
    \item Most of the pictures used to make the meme can be re-edited and a new funny comment can be added to it.
\end{enumerate}

\noindent
\textbf{Note:} In points 6, 7, and 8, the removal of the entity from the images will affect the meaning. In other words, if the entity is removed, then the meaning will not be complete. This is what we mean by constructing meaning. 

% Examples of memes can be found in this guideline.

\subsubsection{Defining Propaganda:}
Propaganda is any communication that deliberately misrepresents symbols and/or entities, appealing to emotions and prejudices while bypassing rational thought, to influence its audience toward a specific goal. Memes are created to be humorous; therefore, it is natural that they lack rational discussion. Instead, they use content to appeal to emotions and prejudices. For our task, we defined the following four categories and annotated the images accordingly.
% For each image, you will choose one of the following categories:  

\begin{figure*}[]
    \centering
    \includegraphics[scale=0.45]{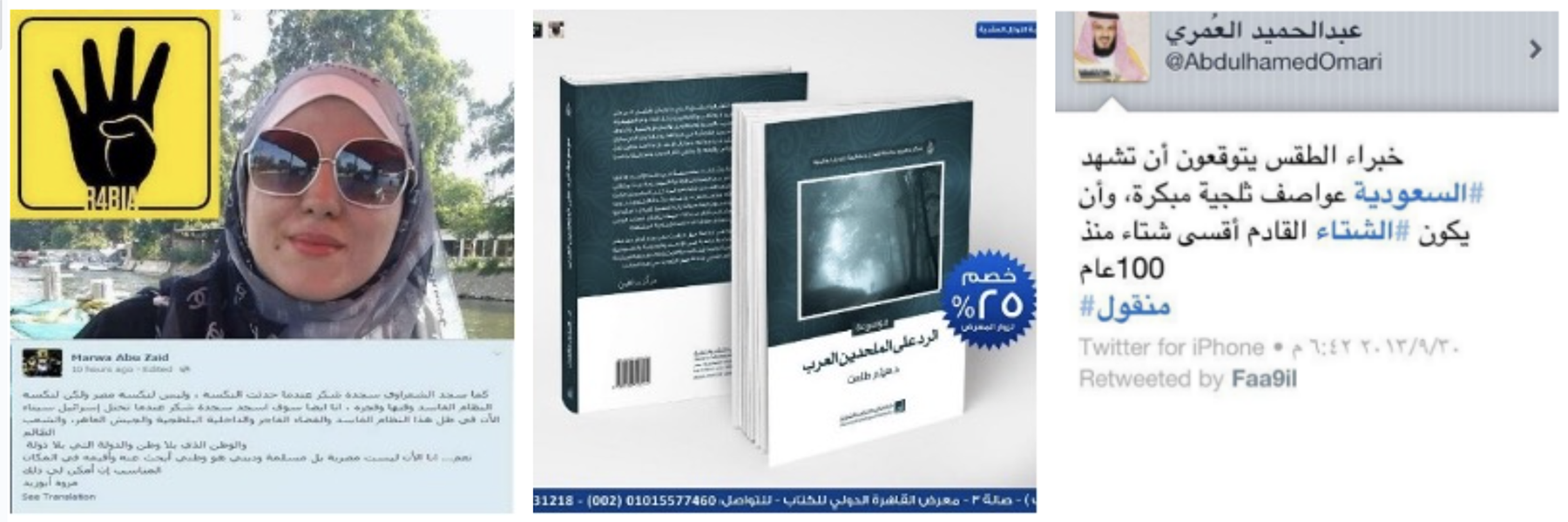}
    \caption{Examples of images labeled as \textit{not-meme}.}
    \label{fig:example_not_meme}
\end{figure*}

\paragraph{(1) Not-Meme:} For images that do not follow the definition of a meme, examples of images labeled as ``not-meme'' are shown in Figure \ref{fig:example_not_meme}.

\begin{figure*}[]
    \centering
    \includegraphics[scale=0.45]{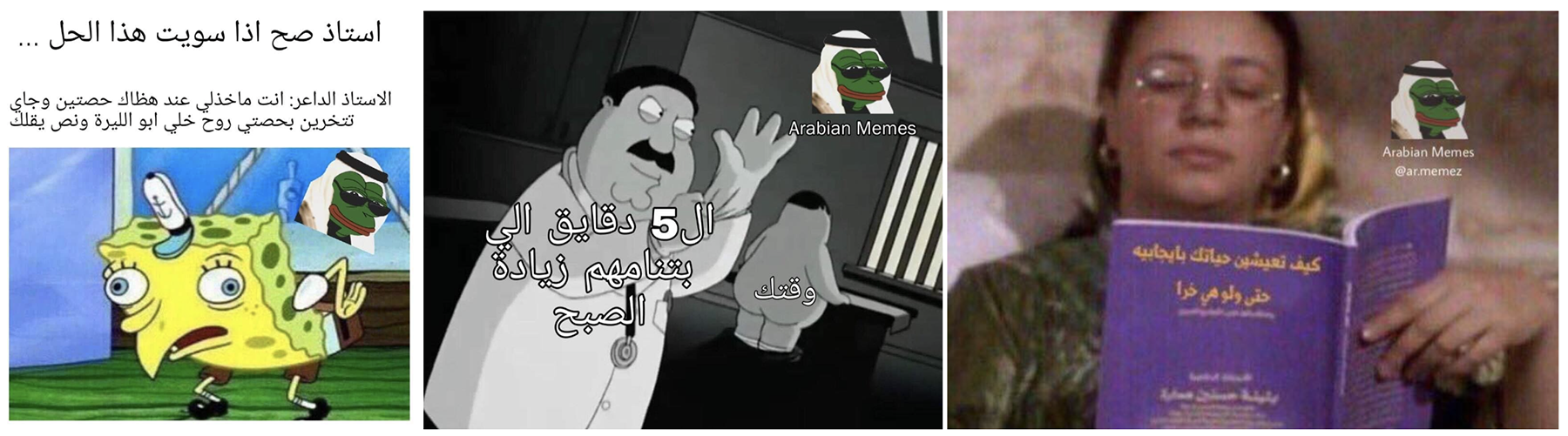}
    \caption{Examples of images labeled as \textit{other}.}
    \label{fig:example_other}
\end{figure*}

\paragraph{(2) Other:} For images that can be defined as memes but fall under any of the criteria listed below. Examples of images labeled as ``not-meme'' are shown in Figure \ref{fig:example_other}.
The criteria for "Other":
\begin{enumerate}
\item Memes that rely on nudity and offensive content, unless the target of the offense is a famous, political, or religious entity.
\item Memes that rely on numbers or figures to construct meaning.
\item Memes that show explicit nudity.
\item Memes that explicitly use offensive words.
\item Memes that are in a different language (not Arabic).
\item Memes that you could not understand due to the dialect it was written in, poor font size, or for any other reason.
\end{enumerate}

\noindent
\textbf{Note:} Memes might contain words that have an implicitly offensive meaning, or uses offensive words may be aimed at social, religious, or political groups. In these cases, the meme does not fall under this criterion. 

\begin{figure*}[h]
    \centering
    \includegraphics[scale=0.52]{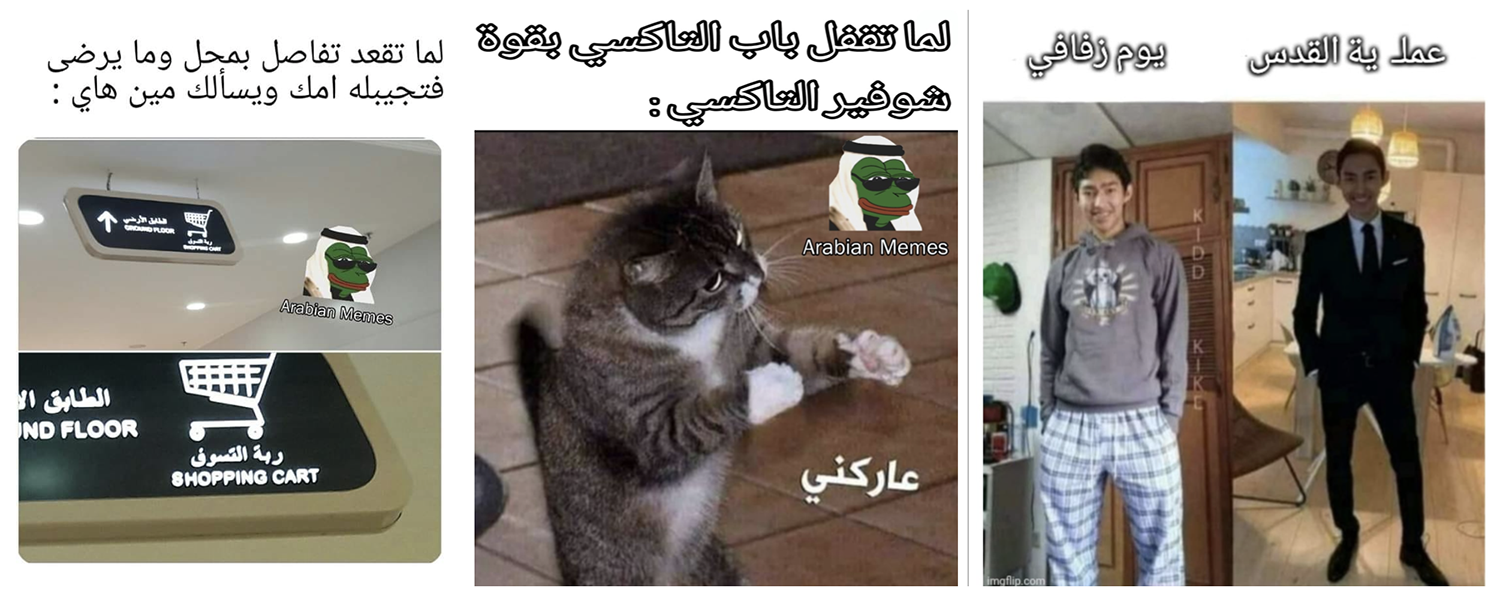}
    \caption{Examples of images labeled as \textit{not propaganda}.}
    \label{fig:example_not_propaganda}
\end{figure*}

\paragraph{(3) Not Propaganda:} For memes that follow the definition of memes but do not contain any propaganda techniques, examples of images labeled as ``not propagandistic'' are shown in Figure \ref{fig:example_not_propaganda}.

\begin{figure*}[h]
    \centering
    \includegraphics[scale=0.52]{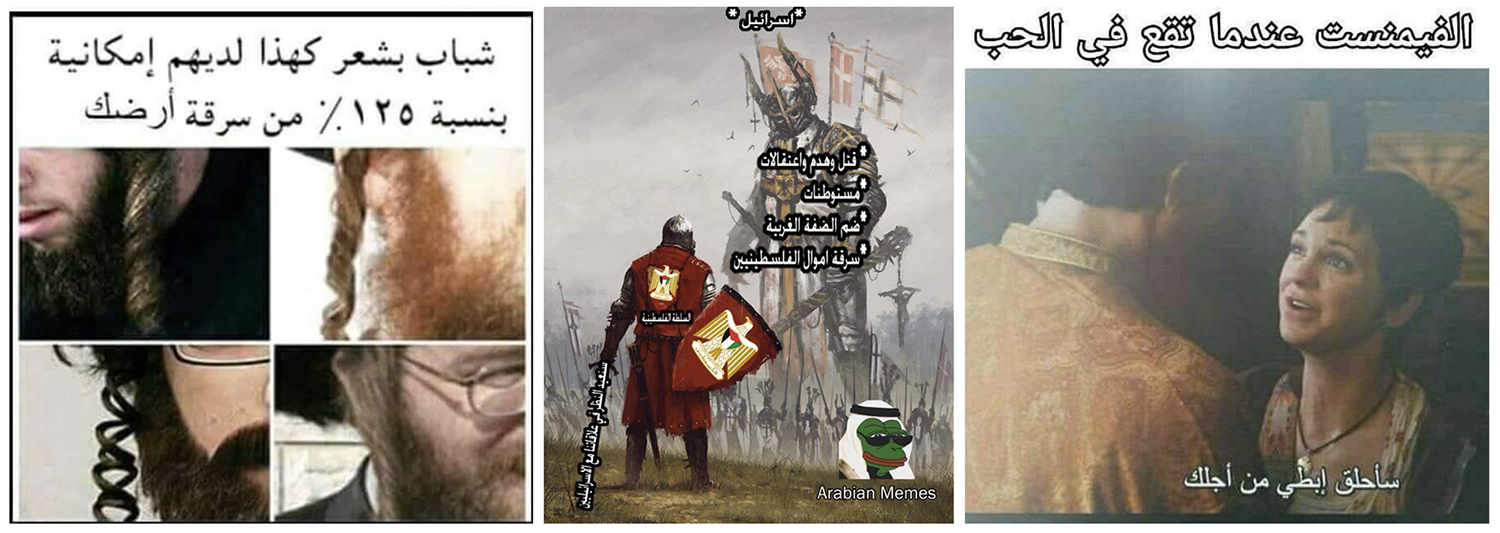}
    \caption{Examples of images labeled as \textit{propaganda}.}
    \label{fig:example_propaganda}
\end{figure*}

\paragraph{(4) Propaganda:} For memes that follow the definition of memes and contain propaganda techniques, examples of images labeled as ``propagandistic'' are shown in Figure~\ref{fig:example_propaganda}.

\subsection{Text Editing}
\label{ssec:app_instructions_text_editing}

The task is to edit the text to match the text shown in the image. The interface will show the picture, alongside the text that is in it. The text was extracted automatically, so it might contain errors. It might not reflect all that is sees in the picture. Some important guidelines to follow for editing the text are listed below: 

\begin{enumerate}
    \item Each part that is a standalone sentence and makes complete meaning should be written as one line. 
    \item Punctuation marks are considered a part of the text. They need to be edited/added.
    \item If the text is in columns, put first all the text of the first column, then all the text of the next column. This task will specifically address memes in Arabic, so the first column should be considered from the right. However, this is not a rule, and memes might change this orientation, so it is up to the annotator to decide the order based on their understanding. 
    \item Rearrange the text so that there is one sentence per line, if possible.
    \item If there are separate blocks of text in different locations of the image, start a new line from each block. 
    \item Leave a blank between two blocks of text if they were shown in two different locations on the picture.
    \item Items that should be excluded from the text: 
    \begin{itemize}
        \item Usernames and social media account names (if visible in the image).
        \item Websites, logos, and any text that is not a part of the meme, so that removing that part does not affect the meaning of the meme.
        \item Any text that is hidden and is hard to read. 
    \end{itemize}
    \item In special cases, a logo can be used in the meme to create meaning. In this case, add the text of the logo to the edited text, if needed.    
\end{enumerate}

\paragraph{Example 1:}
Figure \ref{fig:example_propaganda_text_editing} shows an example of a meme, for editing the text that can be viewed it, the following points are important: 

\begin{figure}[]
    \centering
    \includegraphics[scale=0.35]{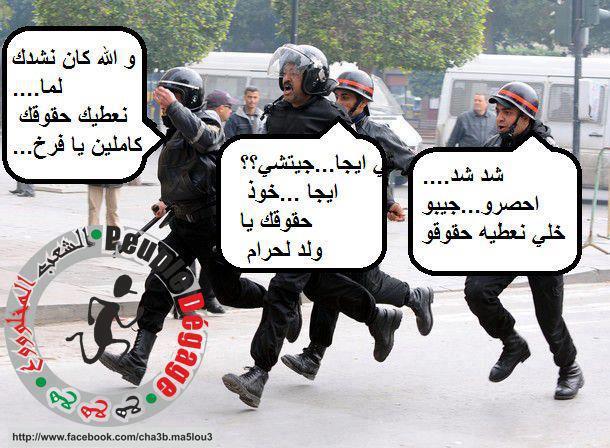}
    \caption{An example of a meme for editing text.}
    \label{fig:example_propaganda_text_editing}
\end{figure}
\begin{itemize}
    \item Each dialog box is one sentence
    \item Start a new line for each box (each box is a different block of text)
    \item Remove any elements that are not part of the meaning: account name and location
    \item Add or modify punctuation to suit what is presented in the text
    \item Text after modification (text translated to EN and read from the first speech bubble from right):
    \begin{verbatim}
Get him ... Get him... corner him... 
get him so we can give him his rights
come... aren't you coming?? 
come...take your rights you son of 
a bastard 
Wallah we gonna get you till... 
we give you all your rights 
you chick ....
\end{verbatim}
\end{itemize}

\begin{figure}[]
    \centering
    \includegraphics[scale=0.28]{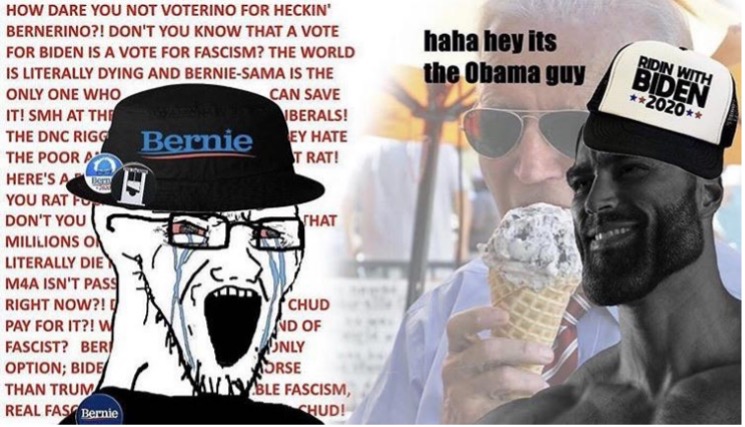}
    \caption{An example of a meme for editing text.}
    \label{fig:example_propaganda_text_editing2}
\end{figure}

\paragraph{Example 2:}
Figure \ref{fig:example_propaganda_text_editing2} shows another example, for which the following points are important. 

\begin{itemize}
    \item Text written in red is difficult to understand and read, so it should not be included in the text.
    \item The text written on the hat and the text in black are each a different block of text. Start a line for each of them and leave a space for each new line.
    \item This example is for illustrative purposes only, and ``memes'' in English will not be shown in this task.
    \item Text after modification:
    \begin{verbatim}    
Bernie
Riding with Biden **2020**
Haha hey its the Obama guy
\end{verbatim}
\end{itemize}

\section{Annotation Platform}
\label{sec:app_annotation_platform}

\subsection{Meme Categorization Task}
\begin{figure*}[h]
    \centering
    \includegraphics[scale=0.8]{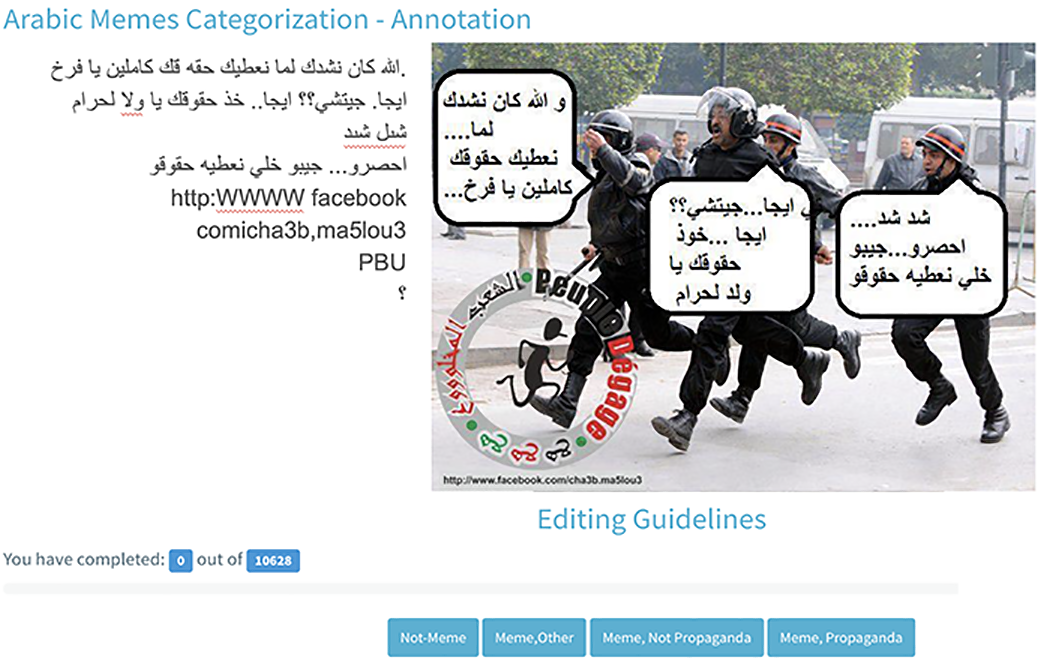}
    \caption{A screenshot of the annotation platform for the \textit{meme categorization} task.}
    \label{fig:annotation_platform_meme_cat}
\end{figure*}

In Figure \ref{fig:annotation_platform_meme_cat}, we provide a screenshot of the annotation platform for the meme categorization task. As shown in the figure, the platform displays the meme itself on the right, the extracted text on the left, a link to the annotation guidelines, and labels with buttons at the bottom for selecting a category for the meme. The task of the annotator was to label the meme as one of the below categories, according to the definitions detailed in the guideline (see Section \ref{sec:annotation_instruct_english}). To facilitate the work of annotators in the annotation process, we used the keywords `meme' along with the labels `other', `propaganda', and `not-propaganda'.
\begin{itemize}
    \item Not Meme 
    \item Meme, Other 
    \item Meme, Not Propaganda 
    \item Meme, Propaganda 
\end{itemize}

\begin{figure*}[h]
    \centering
    \includegraphics[scale=0.25]{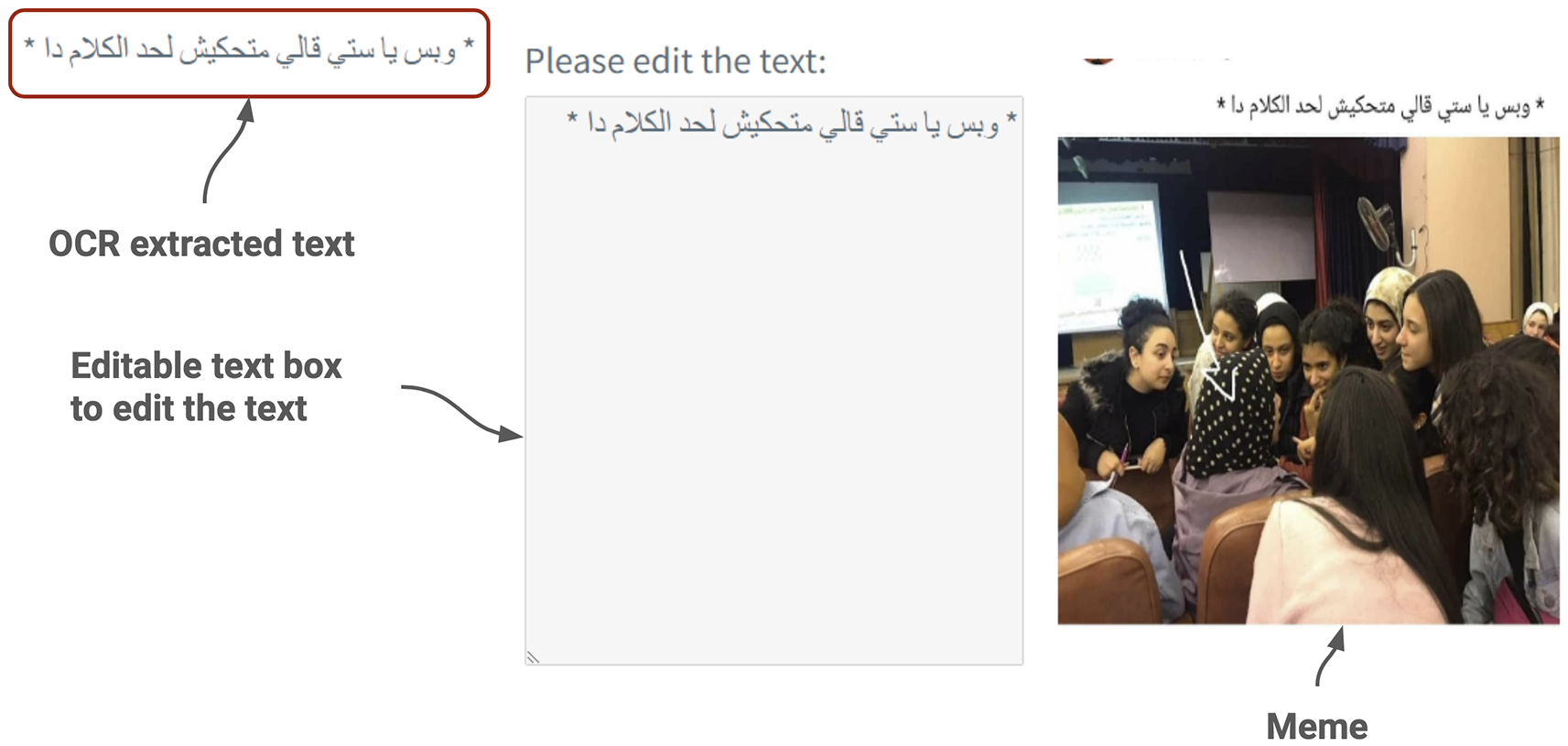}
    \caption{An screenshot of the annotation platform for the \textit{text editing}.}
    \label{fig:annotation_platform_text_editing}
\end{figure*}

% The interface will show you a picture taken from the internet; you are asked to perform the following:

% \begin{enumerate}
%     \item Categorize it under any of the following labels: Not Meme 

% \item Edit the text that was extracted from the picture, however, edit it only for the pictures that you categorized as ``Propaganda'' or ``Not Propaganda''.
% \end{enumerate}
Given that the memes we collected were from different social media platforms, they may contain offensive content. Therefore, we added a note that \textit{some pictures may contain offensive content, and that we apologize for any inconvenience that such content may cause. We appreciate your contribution to this project which will minimize the spread of such harmful content on the internet}. 
% \begin{enumerate}
%     \item Start by studying if the picture shown is a ``meme'' or not. If the picture is not a meme, choose ``Not Meme'', then choose ``Submit'', and the next picture will load.
%     \item If the picture is a “meme”, decide if it falls under ``Other'', then choose ``Submit'', and the next picture will load.
%     \item If the picture does not fall under ``Other'', then choose for it any of the other two labels based on your understanding of the meaning of the meme, then edit the text.
% \end{enumerate}

% \textbf{Some Important Information:} 
To further guide the annotation process, we asked the annotators to follow the following steps. 
\begin{enumerate}
    \item Begin by determining whether the image presented is a ``meme''. If the image is not a meme, select ``Not Meme'', then click ``Submit''. The next image will then be loaded.
    \item If the image is a ``meme'', assess whether it falls under the category of ``Other''. If so, select ``Other'', then click ``Submit''. The next image will then be loaded.
    \item If the image does not fall under the category of ``Other'', choose one of the remaining two labels based on your interpretation of the meme's content. After selecting the appropriate label, edit the text as needed.
\end{enumerate}

\subsection{Text Editing Task}
In this phase, the task was to edit the text based on the guidelines discussed in Section \ref{ssec:app_instructions_text_editing}. In Figure \ref{fig:annotation_platform_text_editing}, we provide a screenshot demonstrating the text extracted from OCR, an editable text box, and the original meme. The task was to edit the text to match it with the original meme.

\section{Arabic Annotation Guideline}
\label{app:arabic_guideline}

\subsection{Meme Categorization}
\subsubsection{Definition of Meme}
\begin{RLtext}\noindent\footnotesize "الميم" أو "الميمز" نص يتكون من صورة تمثل خلفية يضاف إليها نصٌ مكتوبٌ كتعليق أو مزحة، ويمكن لصورة الخلفية أن تكون صورة فوتوغرافية أو رسم توضيحي أو لقطة شاشة، ومن الشائع أن يتم وضع النص في أعلى أو أسفل الصورة ولكن ليس دائمًا، ويؤدي الجمع بين الصورة والنص إلى إنشاء رسالة أو نكتة أو تعليق معين من المفترض أن يسهل فهمه، والتأثر به، ويسهل نشره. متى ستكون الصورة المعروضة "ميمًا"؟
\end{RLtext}
\begin{enumerate}
    \item \begin{RLtext}\noindent\footnotesize تجمع بين نص مكتوب وصورة \end{RLtext}
    \item \begin{RLtext}\noindent\footnotesize النص المكتوب هو تعليق فكاهي \end{RLtext}
    \item \begin{RLtext}\noindent\footnotesize لابد للصورة أن تستوفي الشرطين بالإضافة إلى بعض الخصائص المذكورة بالأسفل.\end{RLtext}
    \item \begin{RLtext}\noindent\footnotesize التعديل على الصورة المستخدمة في الميم.\end{RLtext}
    \item \begin{RLtext}\noindent\footnotesize إضافة نص مكتوب على أماكن متفرقة على الصورة.\end{RLtext}
    \item \begin{RLtext}\noindent\footnotesizeيكثر استخدام صور تعبيرية، أي أنه في وجود كيان (إنسان، حيوان، شخصيات خيالية) بوجه في الصورة، يستخدم التعبيرات المرسومة على الوجه لإنشاء معنى مع التعليق المكتوب.\end{RLtext}
    \item \begin{RLtext}\noindent\footnotesize قد يكون في الصورة كيان يقوم بفعل ما، فيستخدم الفعل لإنشاء معنى مع التعليق المكتوب.\end{RLtext}
    \item \begin{RLtext}\noindent\footnotesize قد يكون في الصورة كيان (منظمة أو شخص)، فيستخدم ما يمثله ذلك الكيان لإنشاء معنى مع التعليق المكتوب.\end{RLtext}
    \item \begin{RLtext}\noindent\footnotesize يكثر استخدام صور وحوارات مأخوذة من أفلام ومسلسلات وإضافة تعليق عليها.\end{RLtext}
    \item \begin{RLtext}\noindent\footnotesize تتميز معظم الصور بأنه يمكن إعادة استخدامها بإضافة تعليق فكاهي آخر.\end{RLtext}
\end{enumerate}
\begin{RLtext}\noindent\footnotesize
ملاحظة: إنشاء المعنى مختلف عن إضافة المعنى، فنقصد في النقاط 6-8  أن عدم وجود تعبير الوجه أو الفعل أو الكيان سيغير المعنى أو سيكون المعنى غير مكتمل. 
\end{RLtext}

\subsubsection{Definition of Propaganda}
\begin{RLtext}
    \noindent\footnotesize
    في حكمك على "الميم" من حيث استخدامها لأسلوب من أساليب البروباغاندا، ستستند على التعريف التالي للبروباغاندا: 
شكل من أشكال التواصل يقوم فيه صاحب الرسالة – متعمدًا – بتشويه الرموز أو إثارة المشاعر أو التحيزات دون اللجوء إلى حجج منطقية وذلك للتأثير على الجمهور ودفعهم نحو هدف معين. 
بما أن "الميم" هو نص في أصله فكاهي، فعلى الأرجح أنك لن تجد فيها حجج منطقية، بل سيستخدم النص (الصورة أو النص المكتوب أو الإثنين) السخرية لتشويه الرموز وإثارة المشاعر.
\end{RLtext}

\begin{figure*}[]
    \centering
    \includegraphics[scale=0.45]{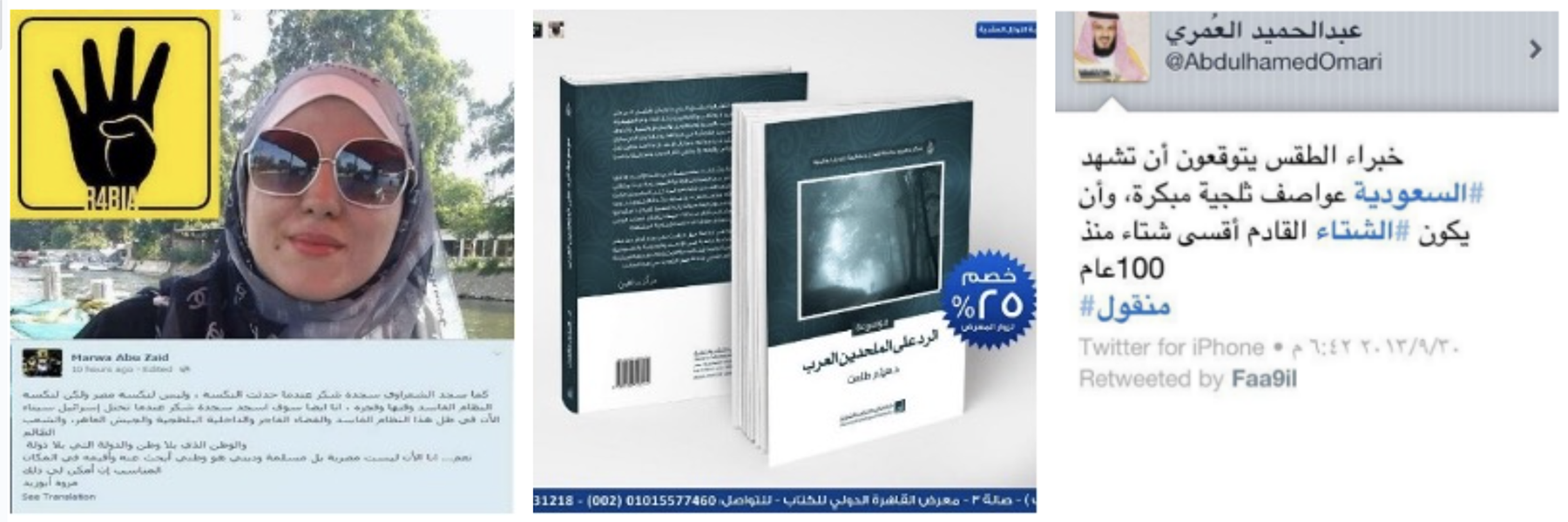}
    \caption{Examples of images labeled as \textit{not-meme}.}
    \label{fig:ar:example_not_meme}
\end{figure*}

\paragraph{(1) Not-Meme: } Figure \ref{fig:ar:example_not_meme} 
\begin{RLtext}
    \noindent\footnotesize يندرج تحت هذا التصنيف الصور التي لا تتبع تعريف "الميمز" المذكور في هذا الدليل. 
\end{RLtext}

\begin{figure*}[]
    \centering
    \includegraphics[scale=0.45]{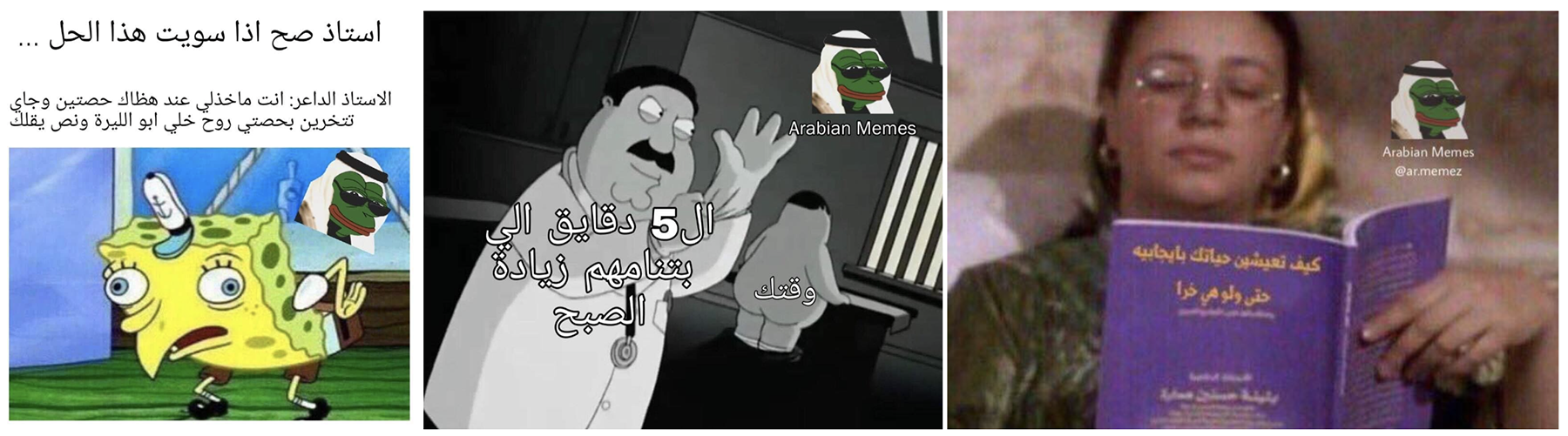}
    \caption{Examples of images labeled as \textit{other}.}
    \label{fig:ar:example_other}
\end{figure*}

\noindent\textbf{(2) Other: } Figure \ref{fig:ar:example_other}
\begin{RLtext}
    \noindent\footnotesize يندرج تحت هذا التصنيف الصور التي تتبع تعريف "الميمز" ولكنها تقع تحت واحدة من المعايير المُحددة في القسم أدناه.
    المعايير والأمثلة: 
\end{RLtext}

\begin{enumerate}
    \item \begin{RLtext}\noindent\footnotesize "الميمز" التي تعتمد على مناظر وشخصيات كرتونية،  ما عدا التي تحتوي على معنى مسيء لشخصيات سياسية أو مشهورة أو جماعات وأحزاب معينة. \end{RLtext}
    \item \begin{RLtext}\noindent\footnotesize "الميمز" التي تعتمد بشكل كلي على أرقام أو مخططات أو رسوم بيانية. \end{RLtext}
    \item \begin{RLtext}\noindent\footnotesize "الميمز" التي تظهر عري صريح. \end{RLtext}
    \item \begin{RLtext}\noindent\footnotesize "الميمز" التي تستخدم كلمات نابية ومنحطة وخادشة للحياء بشكل صريح (مثال: ابن الـ******). \end{RLtext}
    \item \begin{RLtext}\noindent\footnotesize "الميمز" التي تستخدم لغة ثانية غير العربية. (لا يوجد مثال) \end{RLtext}
    \item \begin{RLtext}\noindent\footnotesize "الميمز" التي لم تتمكن من فهمها بسبب اللهجة أو حجم الخط أو لأي سبب آخر.  (لا يوجد مثال) \end{RLtext}
    % \item \begin{RLtext}\noindent\footnotesize \end{RLtext}
    % \item \begin{RLtext}\noindent\footnotesize \end{RLtext}
    % \item \begin{RLtext}\noindent\footnotesize \end{RLtext}
    % \item \begin{RLtext}\noindent\footnotesize \end{RLtext}
\end{enumerate}
\begin{RLtext}\noindent\footnotesize
قد تحتوي "الميمز" على كلمات منحطة لكن بشكل ضمني، أو على شتيمة صريحة غرضها تشويه سمعة أو التقليل (مثال: المنافق، السياسي الفاسد). لا تندرج هذه الكلمات أو العبارات تحت هذا التصنيف، فاختر لها ما يناسبها من التصنيفات الأخرى.
\end{RLtext}

\begin{figure*}[h]
    \centering
    \includegraphics[scale=0.52]{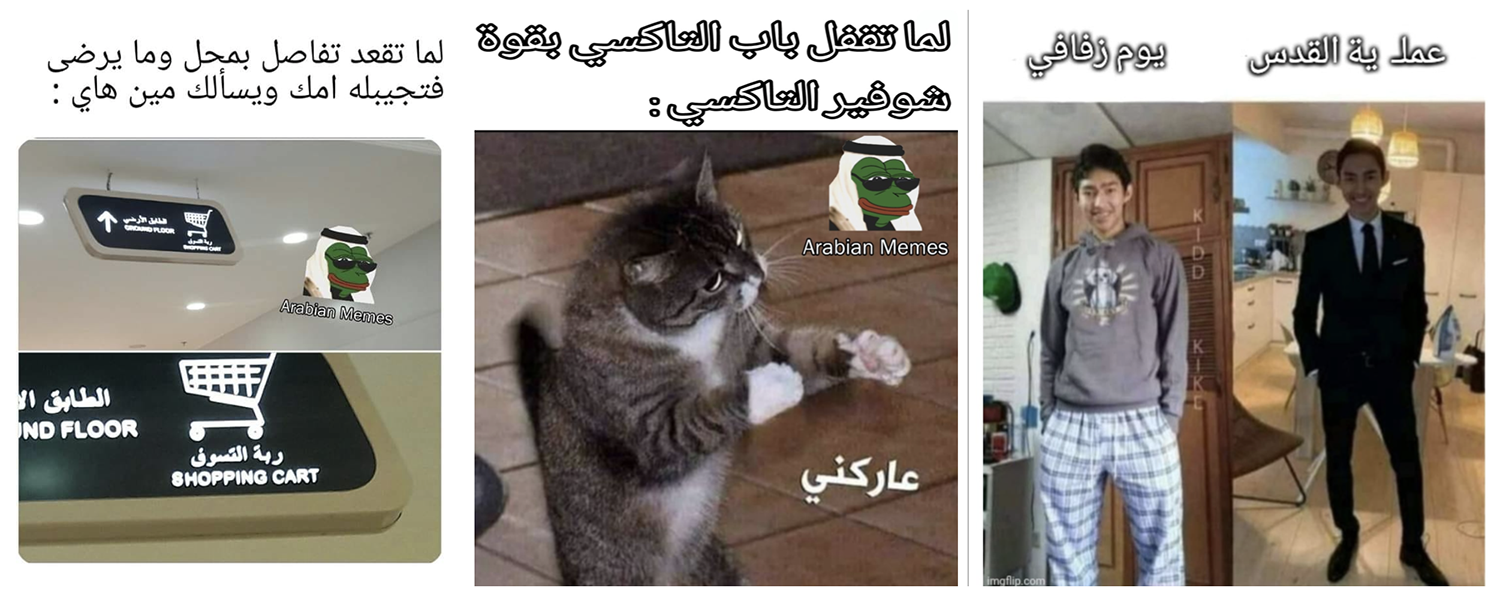}
    \caption{Examples of images labeled as \textit{not propaganda}.}
    \label{fig:ar:example_not_propaganda}
\end{figure*}

\paragraph{(3) Not Propaganda:} Figure \ref{fig:ar:example_not_propaganda}
\begin{RLtext}
    \noindent\footnotesize يندرج تحت هذا التصنيف الصور التي تتبع تعريف "الميمز" ولكنها لا تحتوي على بروباغاندا.
\end{RLtext}

\begin{figure*}[h]
    \centering
    \includegraphics[scale=0.52]{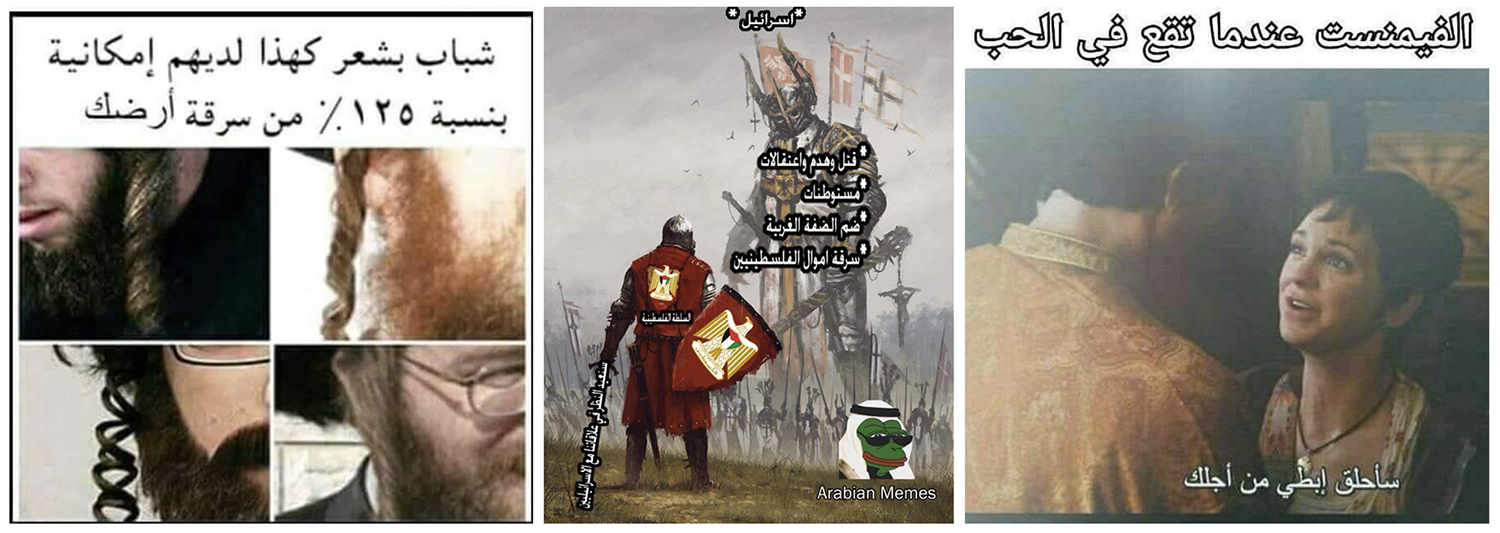}
    \caption{Examples of images labeled as \textit{propaganda}.}
    \label{fig:ar:example_propaganda}
\end{figure*}

\paragraph{(4) Propaganda:} Figure \ref{fig:ar:example_propaganda}
\begin{RLtext}
    \noindent\footnotesize  يندرج تحت هذا التصنيف الصور التي تتبع تعريف "الميمز" وتحتوي على بروباغاندا. 
\end{RLtext}

\subsection{Text Editing}
\begin{RLtext}
    \noindent\footnotesize المطلوب منك في هذه المهمة هو تحرير نصوص عربية، وتعديلها لتطابق المذكور في الصورة المعروضة. ستعرض لك الواجهة صورة مع النص المستخرج منها آليًا، وقد يتضمن النص المستخرج أخطاء أو قد يكون ناقصًا، لكن لا نقصد هنا أخطاء إملائية أو أخطاء نحوية، بل نقصد أن النص لا يطابق المذكور في الصورة.
إليك بعض الإرشادات المهمة:
\end{RLtext}

\begin{enumerate}
    \item \begin{RLtext}\noindent\footnotesize  يجب كتابة كل جملة مستقلة تشكّل معنى في سطر واحد. \end{RLtext}
    \item \begin{RLtext}\noindent\footnotesize علامات الترقيم الواضحة في الصورة تعتبر جزءًا من النص. يجب تحريرها أو إضافتها حسب الحاجة. \end{RLtext}
    \item \begin{RLtext}\noindent\footnotesize إذا كان النص في الصورة معروض في أعمدة، ضع أولاً كل النصوص في العمود الأول، ثم كل النصوص في العمود التالي. ستعمل على نصوص عربية لذا يجب اعتبار العمود الأول من اليمين، لكن الأمر متروك لفهمك لمعنى النص، المهم أن ترتب الجمل لتعطي المعنى المطلوب. \end{RLtext}
    \item \begin{RLtext}\noindent\footnotesize قم بإعادة ترتيب النص بحيث يكون هناك جملة واحدة في كل سطر، إذا كان ذلك ممكنًا.  \end{RLtext}
    \item \begin{RLtext}\noindent\footnotesize إذا كانت هناك كتل أو أجزاء نصية منفصلة في الصورة، ابدأ سطرًا جديدًا لكل كتلة. \end{RLtext}
    \item \begin{RLtext}\noindent\footnotesize اترك فراغًا بين كتلتين من النص إذا عرضا في موضعين مختلفين في الصورة.  \end{RLtext}
    \item \begin{RLtext}\noindent\footnotesize أزل العناصر التالية من النص: \end{RLtext}
        \begin{itemize}
            \item \begin{RLtext}\noindent\footnotesize أسماء المستخدمين وأسماء حسابات وسائل التواصل الاجتماعي لو كانت ظاهرة في الصورة \end{RLtext}
            \item \begin{RLtext}\noindent\footnotesize أي روابط أو نصوص أو شعارات لا تعتبر جزءا من النص الذي يشكل المعنى. \end{RLtext}
            \item \begin{RLtext}\noindent\footnotesize أي نص مخفي ويصعب قراءته. \end{RLtext}
        \end{itemize}
    \item \begin{RLtext}\noindent\footnotesize في حالات خاصة قد تجد أن شعارًا ما استخدم في "الميمز" لإنشاء معنى، عندها أضف نص الشعار إلى النص الذي تحرره. \end{RLtext}
    % \item \begin{RLtext}\noindent\footnotesize  \end{RLtext}
    % \item \begin{RLtext}\noindent\footnotesize  \end{RLtext}
\end{enumerate}

\begin{figure}[]
    \centering
    \includegraphics[scale=0.35]{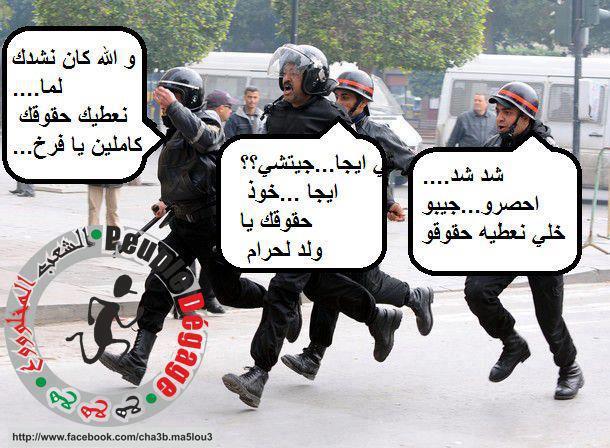}
    \caption{An example of a meme for editing text.}
    \label{fig:ar:example_propaganda_text_editing}
\end{figure}

\paragraph{Example 1: } Figure \ref{fig:ar:example_propaganda_text_editing} 
\begin{RLtext}
    \noindent\footnotesize ملاحظات على تعديل النص في الصورة: 
\end{RLtext}
\begin{enumerate}
    \item \begin{RLtext}
        \noindent\footnotesize كل مربع حوار يعتبر جملة واحدة 
    \end{RLtext}
    \item \begin{RLtext}
        \noindent\footnotesize إبدأ سطرًا جديدًا لكل مربع (كل مربع هو كتلة نصية مختلفة)
    \end{RLtext}
    \item \begin{RLtext}
        \noindent\footnotesize أزل أي عناصر لا تشكل جزءًا من المعنى: اسم الحساب والموقع
    \end{RLtext}
    \item \begin{RLtext}
        \noindent\footnotesize ضف أو عدل علامات الترقيم لتناسب المعروض في النص
    \end{RLtext}
    \item \begin{RLtext}
        \noindent\footnotesize  النص بعد التعديل:
    \end{RLtext}
\end{enumerate}
\begin{RLtext}\noindent\footnotesize
            شد….شد احصرو…جيبو خلي نعطيه حقوقو
            
ايجا…جيتشي؟؟ ايجا …خوذ حقوقك يا ولد لحرام

والله كان نشدك لما…. نعطيك حقوقك كاملين يا فرخ …
        \end{RLtext}

\begin{figure}[]
    \centering
    \includegraphics[scale=0.28]{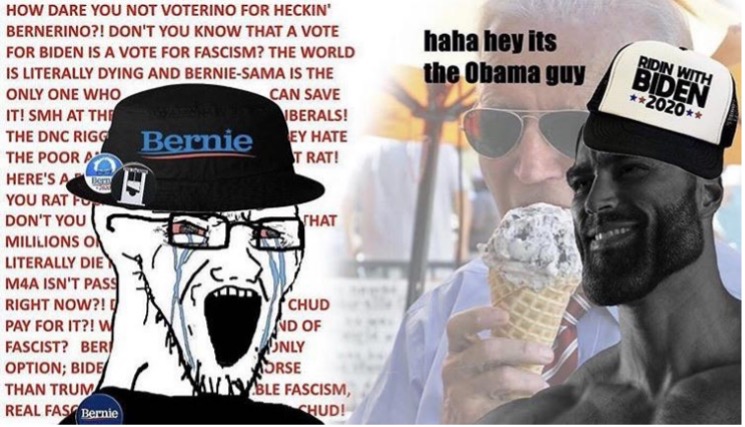}
    \caption{An example of a meme for editing text.}
    \label{fig:ar:example_propaganda_text_editing2}
\end{figure}

\paragraph{Example 2: } Figure \ref{fig:ar:example_propaganda_text_editing2} 
\begin{RLtext}
    \noindent\footnotesize ملاحظات على تعديل النص في الصورة:
\end{RLtext}
\begin{enumerate}
    \item \begin{RLtext}
        \noindent\footnotesize يصعب فهم وقراءة النص المكتوب باللون الأحمر، لذا يجب عدم وضعه من ضمن النص
    \end{RLtext}
    \item \begin{RLtext}
        \noindent\footnotesize النص المكتوب على القبعة والنص الذي بالأسود كل منهم كتلة نصية مختلفة، إبدأ سطرًا لكل منهم واترك فراغًا بس كل سطر جديد.
    \end{RLtext}
    \item \begin{RLtext}
        \noindent\footnotesize هذا المثال لتوضيح حالة فقط، ولن يتم عرض "ميمز" بالإنجليزية في هذه المهمة.
    \end{RLtext}
    \item \begin{RLtext}
        \noindent\footnotesize النص بعد التعديل:
    \end{RLtext}
    \begin{verbatim}
        Bernie 
        Ridin with Biden **2020** 
        Haha hey its the Obama guy
    \end{verbatim}
\end{enumerate}

\begin{figure}[]
    \centering
    \includegraphics[scale=1.2]{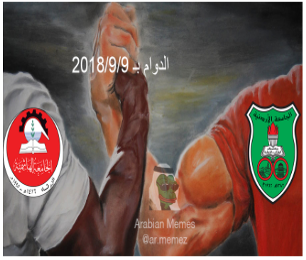}
    \caption{An example of a meme for editing text.}
    \label{fig:ar:example_3}
\end{figure}

\paragraph{Example 3: } Figure \ref{fig:ar:example_3} 
\begin{RLtext}
    \noindent\footnotesize ملاحظات على تعديل النص في الصورة:
\end{RLtext}
\begin{enumerate}
    \item \begin{RLtext}
        \noindent\footnotesize تشكل شعار الجامعة جزءا من المعنى لذا يجب إضافته إلى النص
    \end{RLtext}
    \item \begin{RLtext}
        \noindent\footnotesize نضيف فقط الجزء الذي يسهل قراءته
    \end{RLtext}
    \item \begin{RLtext}
        \noindent\footnotesize النص بعد التعديل:
    \end{RLtext}
    \begin{RLtext}
        \noindent\footnotesize الدوام بـ9$\backslash$9$\backslash$2018

الجامعة الأردنية

الجامعة الهاشمية
    \end{RLtext}
\end{enumerate}

\end{document}